\documentclass[a4paper,12pt]{article}
\usepackage[utf8]{inputenc}
\usepackage[T1]{fontenc}
\usepackage[english]{babel}
\usepackage{natbib}
\usepackage{graphicx}
\usepackage{booktabs}
\usepackage{adjustbox}
\usepackage[a4paper, total={6in, 8in}]{geometry}
\usepackage{amsmath}
\usepackage{rotating,tabularx,siunitx,caption}
\usepackage{setspace}
\usepackage{xltabular}
\usepackage{eurosym}
\usepackage{comment}
\usepackage{xtab}
\usepackage{url}
\usepackage{hyperref}
\usepackage{float}  
\usepackage{rotating}
\usepackage{threeparttable} 
\usepackage{longtable}
\usepackage{multirow}
\usepackage{authblk}
\usepackage{graphicx}

\begin{document} 

\title{Discourse vs emissions: Analysis of corporate narratives, symbolic practices, and mimicry through LLMs}

\author[1, 3, 4 , 5]{Bertrand Kian Hassani}
\author[1]{Yacoub Bahini}
\author[2]{Rizwan Mushtaq}

\renewcommand\Affilfont{\footnotesize} 
\affil[1]{QUANT AI Lab, C. de Arturo Soria, 122, 28043 Madrid, Spain}
\affil[2]{EDC Paris Business School, Paris, France}
\affil[3]{Department of Computer Science, University College London, Gower Street, London WC1E 6EA, UK}
\affil[4]{Centre d'Économie de la Sorbonne (CES), Maison des Sciences Économiques (MSE), Université Paris 1 Panthéon-Sorbonne, 106-112 Boulevard de l'Hôpital, 75013 Paris, France}
\affil[5]{Institut Louis Bachelier, Palais Brongniart, 28 Place de la Bourse, 75002 Paris, France}

\date{\today}
\vspace{-2em}
\maketitle

\vspace{-2em}
\begin{abstract}

Climate change has increased demands for transparent and comparable corporate climate disclosures, yet imitation and symbolic reporting often undermine their value. This paper develops a multidimensional framework to assess disclosure maturity among 828 U.S.-listed firms using large language models (LLMs) fine-tuned for climate communication. Four classifiers—sentiment, commitment, specificity, and target ambition—extract narrative indicators from sustainability and annual reports, which are linked to firm attributes such as emissions, market capitalization, and sector. Analyses reveal three insights: (1) risk-focused narratives often align with explicit commitments, but quantitative targets (e.g., net-zero pledges) remain decoupled from tone; (2) larger and higher-emitting firms disclose more commitments and actions than peers, though inconsistently with quantitative targets; and (3) widespread similarity in disclosure styles suggests mimetic behavior, reducing differentiation and decision-usefulness. These results highlight the value of LLMs for ESG narrative analysis and the need for stronger regulation to connect commitments with verifiable transition strategies.

\end{abstract}

{\small 

~\\
\vspace{0em}
}

\newpage

\section{Introduction}

Climate change has emerged as one of the most pressing global challenges, driving governments, regulators, investors, and civil society to demand greater transparency from companies regarding their environmental strategies and impacts. Corporate sustainability reports, particularly those focusing on climate change, have become central to how firms communicate their risk management practices, mitigation strategies, and long-term transition plans. International frameworks such as the Task Force on Climate-related Financial Disclosures (TCFD), the Global Reporting Initiative (GRI), and, more recently, the European Sustainability Reporting Standards (ESRS) have increased expectations for comprehensive and comparable climate-related disclosure. In parallel, a successful transition to a decarbonized economy requires not only the existence of commitments and emission reduction targets but also credible, specific, and actionable narratives that align with firms’ actual performance and sectoral context.\\

\noindent Recent empirical evidence have demonstrated that corporate sustainability disclosure practices often involve imitation of peer firms, rather than unique, performance-driven communication (\cite{huang2025imitation}, \cite{huang2025unmasking}). Studies grounded in institutional theory document that firms benchmark their ESG disclosures against industry norms, sometimes prioritizing conformity and reputational management over substantive differentiation \citep{AERTS2006299, huang2025imitation}. This imitation behavior can reduce the informational value of climate disclosure if narratives become standardized, vague, or disconnected from operational realities. At the same time, signaling theory suggests that some firms adopt credible, verifiable commitments (e.g., Science-Based Targets, TCFD alignment) to differentiate themselves from peers and demonstrate long-term climate resilience \citep{bolton2021mandatory, flammer2021shareholder}.\\

\noindent Advances in artificial intelligence such as natural language processing (NLP) and large language models (LLMs) now enable more sophisticated measurement of disclosure quality, capturing narrative tone, specificity, and target ambition beyond traditional metrics such as report length or keyword frequency \citep{bingler2023cheaptalk, zou2025esgreveal}. For instance, \citet{bingler2023cheaptalk} leverage transformer-based models to analyze corporate climate commitments and identify patterns of ``cheap talk'' across sectors and over time. While their focus lies on commitment content and temporal dynamics, their study does not explicitly consider narrative inconsistencies or imitation behaviors, which other research suggests may mask weak internal governance or limited operational alignment with stated climate goals \citep{huang2025imitation}. Therefore, this study attempts to fill this gap by addressing following questions, (i) do the claims in corporate sustainability report aligns with the commitments and objectives ? (ii) whether corporate climate disclosure follows mimetic patterns ? (iii) do the firm specific indicators such as CO2 emissions, firm size, market capitalization and sector of activity derive corporate climate narratives ?  \\

\noindent Our study contributes to three main areas of academic debate: first, the interplay between credible and symbolic communication in climate narratives. Using four specialized LLM classifiers (sentiment, commitment, specificity, and target ambition), originally fine-tuned by \citet{bingler2023cheaptalk} applied in this study to construct corporate climate narrative indicators. Second, the role of firm size and sector in shaping disclosure quality and consistency. Our framework examines how these narratives relate to firm characteristics such as CO$_2$ emission levels, market capitalization, workforce size and sector. Third, the potential emergence of sector-wide mimetic behaviors that reduce informational value for investors and stakeholders. In doing so, we provide novel insights into whether firms truly ``talk their walk'' and how their climate narratives align with the broader goal of supporting a successful transition to a decarbonized economy.   This enables us to detect potential inconsistencies, for instance when net-zero targets are disclosed without supporting commitments, or when ambitious climate objectives are accompanied by vague narratives. Such patterns may point to symbolic disclosure, greenwashing, or mimicking behaviors rather than substantive engagement \citep{bebbington2001sustainable, bini2016put}.\\

\noindent \noindent Our results reveal important patterns underlying corporate climate narratives. First, risk-oriented narratives are generally accompanied by more explicit and specific commitments, indicating that firms using a cautious tone tend to provide clearer action plans. However, quantitative targets such as net-zero pledges often appear disconnected from narrative tone, suggesting a decoupling between rhetoric and measurable objectives. Second, larger and higher-emitting firms show slightly higher disclosure maturity, with more frequent commitments and specific actions, yet these remain inconsistently aligned with quantitative targets—pointing to symbolic practices. Third, the strong convergence of disclosure styles across sectors and firm sizes indicates widespread mimetic behavior, which reduces differentiation and weakens the informational value of climate disclosures for investors and stakeholders. These findings reinforce our theoretical contribution by linking narrative tone, commitments, and firm attributes to the broader debate on credibility versus symbolism in corporate climate reporting.

\section{Literature review and hypotheses}

\subsection{Theoretical background}

Research on corporate climate-related disclosure often builds on two main theoretical perspectives: the \textit{voluntary disclosure theory} and the \textit{legitimacy theory} (\cite{PARK2023135203, deegan2002introduction}).  
The voluntary disclosure theory argues that firms with superior environmental performance disclose more and with higher quality to reduce information asymmetry, lower capital costs, and signal credible long-term risk management strategies (\cite{VERRECCHIA1983179, CLARKSON2008303}).  
Conversely, legitimacy theory posits that poor performers disclose extensively—often using vague or symbolic communication—to maintain social acceptance or repair reputation, even in the absence of substantive change \cite{CHO2007639, Suchman1995}.  
Empirical results are mixed: some studies support voluntary disclosure motives (e.g., \citealt{LU2021101264}), while others confirm legitimacy-driven disclosure behavior (e.g., \citealt{Jiang2022, SILVA2021125962}).\\  

\noindent Recent evidence suggests that these theoretical perspectives can be complementary rather than contradictory.  
For example, firms with strong environmental performance may disclose extensively to signal credibility and reduce information asymmetry (voluntary disclosure logic), while at the same time, even these firms may resort to symbolic or selective reporting in areas where performance is weaker (legitimacy logic).  
Similarly, poor performers may engage in extensive disclosure to maintain legitimacy, but they can also strategically highlight genuine improvements to lower financing costs and attract investors.  
This overlap illustrates how both theories may operate simultaneously within the same firm, depending on the issue, the audience, or the reporting context \citep{PARK2023135203, bingler2023cheaptalk}.  
 \\

\noindent Beyond the voluntary vs. legitimacy debate, disclosure quality and credibility are shaped by additional channels:  
(i) credibility signaling via specific commitments (e.g., SBTi participation, TCFD compliance) \citep{bolton2021mandatory, bingler2022cheap},  
(ii) investor engagement and ownership pressure \citep{ilhan2023climate, flammer2021shareholder}, and  
(iii) reputational risk management following environmental controversies \citep{bebbington2001sustainable, bini2016put}.  
\noindent These factors can incentivize symbolic communication, mimicry of industry leaders, or substantive transparency improvements depending on context and firm characteristics \citep{baldini2018role}.

\subsection{Gap in the literature}\label{gap}

Much of the early research on climate disclosure has relied on basic proxies such as report length or indicator counts. More recent contributions apply natural language processing (NLP) techniques to capture narrative aspects of sustainability reporting. Some adopt relatively simple, dictionary-based approaches to measure tone or framing \citep[e.g.,][]{huang2025unmasking}, while others employ deep learning models to detect subtler signals of credibility or greenwashing \citep{bingler2023cheaptalk, zou2025esgreveal}.  \noindent Among these, \citet{bingler2023cheaptalk} fine-tune ClimateBERT classifiers to identify climate-related ``cheap talk'' in corporate disclosures of MSCI World firms. Their approach provides valuable insights by showing how voluntary disclosures are often associated with more \textit{cheap talk}, how targeted climate engagement reduces it, and how cheap talk correlates with negative news coverage, higher emissions growth, and greater transition and reputational risk exposure. \\

\noindent By contrast, the present study leverages the same ClimateBERT classifiers (sentiment, commitment, specificity, and target ambition) but constructs a distinct set of indicators. Instead of using credibility measures, we apply paragraph-level classification combined with threshold-based aggregation rules to assign report-level orientations (e.g., risk vs. opportunity, commitment vs. no commitment, specific vs. general, reduction vs. net-zero). This methodology -  detailed in section \ref{methodo} - enables a systematic evaluation of narrative content. 

\noindent Furthermore, our approach links these narrative indicators to firm-level characteristics (e.g., CO$_2$ emissions, market capitalization, workforce size) and sectoral differences. This cross-sectional design complements prior longitudinal work by revealing how disclosure maturity varies across firms and contexts, and by detecting inconsistencies between narrative tone, stated commitments, and quantitative performance. In doing so, it contributes to a deeper understanding of symbolic disclosure, imitation behavior, and greenwashing in the transition to a low-carbon economy.

\subsection{Hypothesis}

Based on prior literature on voluntary disclosure, legitimacy theory, and disclosure dynamics across sectors, we formulate the following hypotheses.\\ 

H1: Firms with higher greenhouse gas emissions are more likely to issue generic, low-specificity climate commitments or omit explicit commitments altogether, reflecting symbolic disclosure aimed at legitimacy management.

H2: Firms with lower emissions are more likely to provide specific, verifiable climate commitments and ambitious net-zero targets, signaling proactive risk management and long-term transition strategies.

H3: Inconsistencies between narrative tone (e.g., opportunity-oriented vs.\ risk-oriented framing) and concrete targets (e.g., net-zero pledges without substantive commitments) indicate potential greenwashing or mimicking behavior.

H4: Larger firms, whether measured by market capitalization or workforce size, are more likely to adopt explicit commitments and higher-quality disclosures due to stronger stakeholder and regulatory scrutiny.

H5: Sectoral differences shape disclosure patterns: high-transition-risk sectors (e.g., energy, utilities) face stronger legitimacy pressures leading to symbolic or mimicked disclosures, whereas low-emission sectors (e.g., finance, technology) tend to exhibit more opportunity-oriented narratives with varying specificity.

H6: Firms, particularly within the same sector, tend to mimic the disclosure style and narrative framing of their peers, reflecting institutional pressures and peer benchmarking.

\bigskip

\noindent First, we expect that \textit{firms with higher greenhouse gas emissions are more likely to issue generic climate commitments (low specificity) or omit explicit commitments altogether} (H1).  
Such patterns are consistent with legitimacy-driven symbolic disclosure, in which communication serves primarily to manage societal expectations rather than to convey substantive climate strategies (\citep{PARK2023135203}).

\noindent Second, we hypothesize that \textit{firms with lower emissions are more likely to provide specific, verifiable climate commitments and ambitious net-zero targets} (H2).  
This reflects voluntary disclosure incentives, where firms with better environmental performance seek to credibly signal their proactive risk management and long-term transition strategy (\citep{CLARKSON2008303}).  

\noindent Third, we posit that \textit{inconsistencies between narrative tone} (e.g., opportunity-oriented framing versus risk-oriented framing) and concrete targets (e.g., net-zero pledges without underlying commitments or specificity) may indicate potential greenwashing or mimicking behavior (H3).  
Such inconsistencies suggest that some firms “talk their walk” in ways that imitate peers’ communication patterns without corresponding internal alignment (\citep{bingler2023cheaptalk, huang2025imitation}).  

\noindent Fourth, we expect that firm size influences disclosure quality: \textit{larger firms are more likely to adopt explicit commitments and detailed disclosures than smaller firms} due to heightened stakeholder and regulatory scrutiny (H4).

\noindent Fifth, sectoral differences are expected to emerge (H5).  
Sectors with high transition risks, such as energy or utilities, may face stronger legitimacy pressures leading to symbolic or mimicked disclosures, while sectors with lower direct emissions, such as finance or technology, may exhibit more opportunity-oriented narratives with varying levels of specificity (\cite{manes2022exploring}.

\noindent Finally, we anticipate evidence of mimetic behavior (H6), whereby firms, particularly within the same sector, adopt similar disclosure styles and narrative framings regardless of their actual emissions levels or strategic climate engagement, reflecting institutional pressures and peer benchmarking (\citep{AERTS2006299, huang2025imitation}).

\section{Methodology}\label{methodo}

\subsection{Data sources and rationale}

We compile a cross-sectional dataset of 828 U.S.-listed companies for the reference year 2023, combining three main sources of information. \\
\textbf{Corporate documents:} ESG or sustainability reports are prioritized\footnote{Expected to contain more financially material climate information.}; in their absence, annual reports or other official corporate disclosures are used. For each company, we select the most recent report available at the time of collection, with the majority of documents published in 2023 or 2024, a few in 2025, very few in 2022, and only a handful in 2021. This choice of using a single, most recent document is consistent with our cross-sectional research design and is justified by the relative stability of ESG narratives over short time horizons, which limits the added value of incorporating multiple reports from adjacent years. Reports were collected manually, as automated retrieval tools currently present limitations in accurately identifying the relevant disclosure for each company.  
  \\
\textbf{Emissions data ($CO_2$):} Scopes 1, 2, and 3 carbon dioxide (CO\textsubscript{2}) emissions are sourced from Refinitiv ESG.  \\
\textbf{Firm characteristics:} Market capitalization, workforce size, and sector information are obtained from Yahoo Finance and Eulerpool.

\noindent Using one representative document per company (cross section) is justified by the relative stability of ESG narratives over short time horizons.

\subsection{Methods}

This study leverages large language models (LLMs) to analyze climate-related discourse among U.S.-listed companies. All firms operate under a unified regulatory framework that enforces consistent ESG disclosure requirements.

\subsection{Textual climate disclosure classification}

To analyze corporate climate narratives, we rely on four ClimateBERT models fine-tuned by \citet{bingler2023cheaptalk}\footnote{See Section~\ref{gap} for a brief discussion of \citet{bingler2023cheaptalk}; for further details, refer to the original paper}. The models are trained at the paragraph level and therefore cannot be directly applied to entire reports. These models are available on Hugging Face and are pre-trained for climate-specific text classification to capture different dimensions of climate communication.\footnote{Model links: 
\url{https://huggingface.co/datasets/climatebert/climate_sentiment}, 
\url{https://huggingface.co/climatebert/distilroberta-base-climate-commitment}, 
\url{https://huggingface.co/climatebert/distilroberta-base-climate-specificity}, 
\url{https://huggingface.co/climatebert/netzero-reduction}.}  

\noindent Specifically, the \textit{Sentiment} model classifies climate-related text into \textit{risk}, \textit{neutral}, or \textit{opportunity} narratives, reflecting how companies frame climate change. The \textit{Commitment} model detects whether firms explicitly communicate climate mitigation or adaptation commitments. The \textit{Specificity} model differentiates between general statements and specific, verifiable climate actions, in line with the \citet{ISAB2018} concept of materiality. Finally, the \textit{Quantitative targets} model identifies whether companies disclose explicit emission reduction goals or net-zero pledges.

\subsection{Text extraction and keyword filtering}

To focus the analysis on climate-relevant content, we first identify and extract paragraphs that are likely to contain information on emissions and climate strategy. \citet{bingler2023cheaptalk} developed and fine-tuned a dedicated model to automatically detect such climate-related paragraphs. In our study, we adopt  a simpler and rule-based approach by applying a predefined set of keywords to filter the text. This keyword-based filtering ensures that only relevant sections of corporate documents are retained for subsequent analysis. The keyword list is organized into the following groups:

\noindent
\textbf{Greenhouse gases} (\texttt{co2}, \texttt{co\_2}, \texttt{ghg}, \texttt{greenhouse gas}, \texttt{carbon footprint}), \textbf{Emission scopes} (\texttt{scope 1}, \texttt{scope 2}, \texttt{scope 3}, \texttt{Emission scopes}), \textbf{Targets and neutrality} (\texttt{net zero}, \texttt{carbon neutrality}, \texttt{emission reduction}, \texttt{zero emission}, 
\texttt{overall emissions}, \texttt{cutting emissions}, \texttt{emissions footprint}, \texttt{climate target}), 
\textbf{Strategy and risks} (\texttt{decarbonization}, \texttt{climate strategy}, \texttt{transition risk}, \texttt{carbon intensity}, \texttt{climate change}, 
\texttt{emission from}, \texttt{direct emission})

\noindent For each report, each paragraph is first classified independently by the four ClimateBERT models, then we compute ratios of classified paragraphs and apply thresholds to assign the overall 
narrative orientation at the report level:

\begin{itemize}
    \item \textit{Global sentiment:} we compute the ratio of paragraphs classified as \textit{risk} and as \textit{opportunity}.  
    A report is labeled as \textit{risk} if more than 30\% of paragraphs are risk-oriented 
    and fewer than 30\% are opportunity-oriented. 
    It is labeled as \textit{opportunity} if more than 30\% of paragraphs are opportunity-oriented 
    and fewer than 30\% are risk-oriented. 
    If both exceed these thresholds, the report is labeled as \textit{Global risk\_opportunity}. 
    Otherwise, it is labeled as \textit{neutral}.
    \item \textit{Commitment:} we compute the share of paragraphs explicitly mentioning climate commitments. 
    If this share exceeds 40\%, the report is classified as \textit{commitment}, otherwise as \textit{no commitment}.
    \item \textit{Global Specificity:} we compute the share of paragraphs expressing specific (verifiable) actions. 
    Reports with more than 40\% specific statements are labeled as \textit{specific}, others as \textit{general}.
    \item \textit{Global Net-zero and reduction targets:} we compute the proportion of paragraphs 
    mentioning net-zero pledges and those mentioning reduction targets.  
    If both ratios exceed 30\%, the label is \textit{reduction\_netzero};  
    if only net-zero exceeds 30\%, the label is \textit{reduction};  
    if only reduction exceeds 30\%, the label is \textit{netzero};  
    otherwise, the report is labeled as \textit{no reduction}.
\end{itemize}

\noindent This threshold-based aggregation method provides a more robust narrative classification 
by accounting for the relative importance of each disclosure type in the overall report. Then, the outputs are merged with quantitative indicators (Scope 1, 2 and 3 emissions, market capitalization, workforce size). Firms are then categorized into eight classes for each continuous variable, including emission scopes 1, 2 and 3, market capitalization, and number of employees to facilitate typological analysis.\\
\noindent Sample completeness is preserved across analyses by firm size (market capitalization and number of employees),  sector and scope 1 emissions. However, analyses by emission class (Scopes 2, 3) involve smaller subsets due to missing emission data (11 missing values for scope 2 and 307 for scope 3). 

\subsection{Clustering and taypology analysis}

We perform unsupervised clustering to identify patterns linking climate narratives to emissions. First, narrative and quantitative variables are standardized. Clustering is then conducted using a Gaussian Mixture Model (GMM), which allows for non-spherical clusters and probabilistic membership assignment \citep{azizyan2015efficient, bakshi2020outlier}. The optimal number of clusters is determined via the Bayesian Information Criterion (BIC). Finally, clusters are profiled based on narrative variables (sentiment, commitment, specificity, targets), emission levels, and firm characteristics such as sector and size. This approach enables an integrated assessment of how corporate climate narratives—covering tone, commitments, and targets—relate to actual emission performance and structural firm attributes.

\section{Results}

\subsection{Interrelations in corporate climate communication}


\subsubsection{Correlation patterns}

\begin{figure}[htbp]
    \centering
    \includegraphics[width=0.7\textwidth]{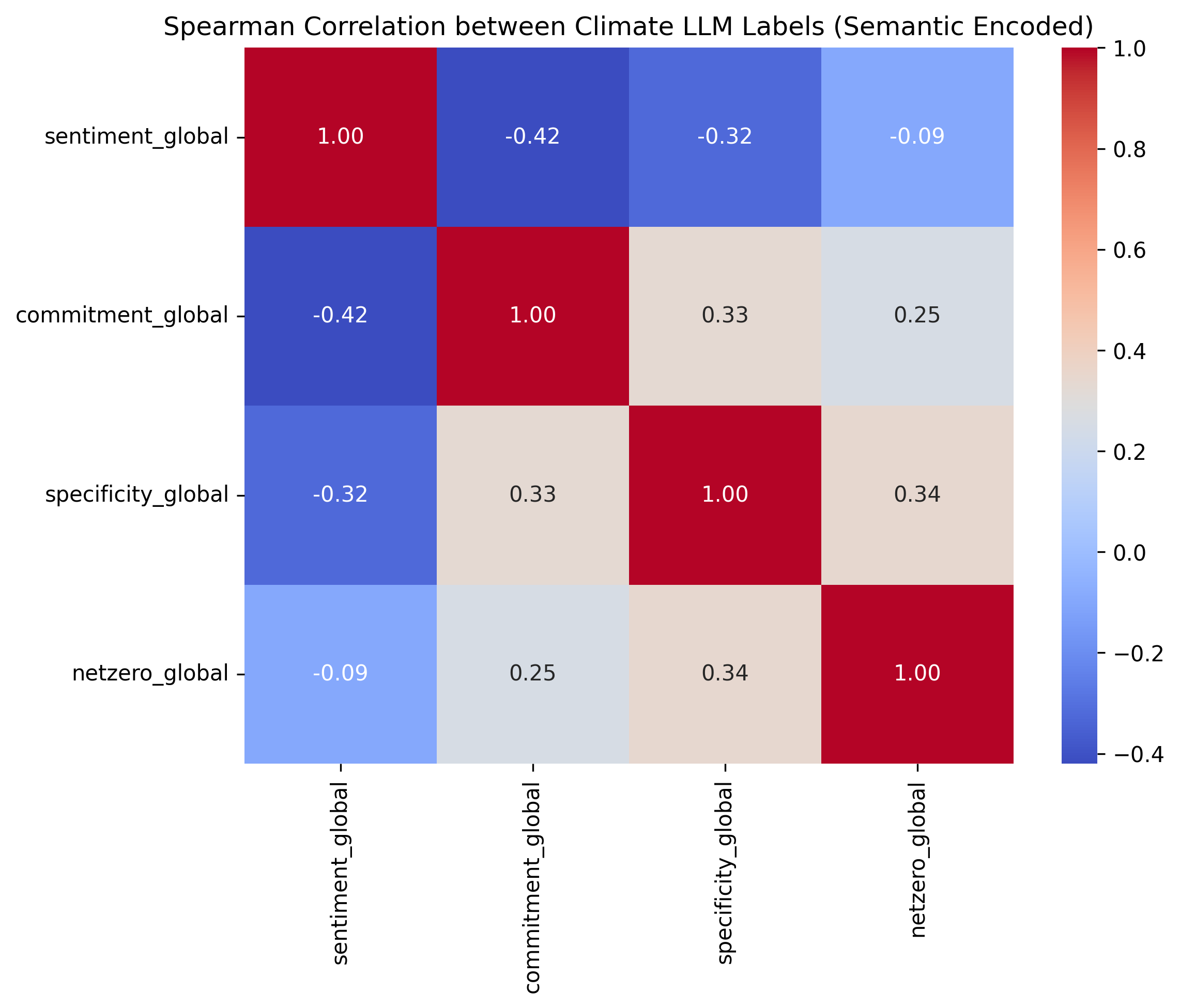}
    \caption{Spearman correlation matrix between Climate LLM labels (semantic encoding). Thresholds used: commitment $>$ 40\%, specificity $>$ 40\%, risk $>$ 30\%, opportunity $>$ 30\%, net-zero $>$ 30\%, reduction $>$ 30\%.}
    \label{fig:spearman-corr}
\end{figure}

Table~\ref{tab:semantic-encoding} defines the semantic order used for categorical LLM labels, enabling an ordinal treatment of sentiment, commitment, specificity, and net‑zero target ambition.  
The resulting Spearman correlation coefficients, $\rho$, are presented in Figure~\ref{fig:spearman-corr}, with corresponding p‑values provided in Table~\ref{tab:spearman-pvalues-compact}.  
The p‑values are very close to zero for all pairs, indicating that correlation coefficients obtained are statistically significant.  
These results provide insight into how different dimensions of climate discourse and commitments interact. 

\paragraph{Sentiment and commitment:} 
A weak negative correlation is observed ($\rho=-0.42$), suggesting that firms with explicit climate commitments tend to adopt a more risk‑oriented tone rather than emphasizing opportunity.  
This indicates that climate‑engaged companies often frame climate change as a risk management issue, likely to justify strategic responses and satisfy regulatory or stakeholder expectations.

\paragraph{Sentiment and specificity:} 
The correlation between sentiment and specificity is also weak and negative ($\rho=-0.32$), showing that firms providing more detailed and actionable climate commitments tend to adopt risk‑oriented narratives rather than neutral or opportunity framing.  
This reinforces the idea that operationalizing climate discourse through specific actions is often accompanied by risk‑focused communication.

\paragraph{Sentiment and net‑zero ambition:} 
The relationship between sentiment and quantitative climate targets is very weak ($\rho=-0.09$), indicating that the tone of climate discourse is largely independent from the decision to set quantitative reduction or net‑zero goals.  
This suggests that narrative orientation (risk, neutral, opportunity) and quantitative target adoption are distinct dimensions of climate communication.

\paragraph{Commitment and specificity:} 
Commitment and specificity are weakly positively correlated ($\rho=0.33$), indicating that firms expressing climate commitments are more likely to articulate them in specific and verifiable terms.  
However, the modest correlation implies that many commitments remain broad or normative, revealing heterogeneity in how firms operationalize their climate pledges.

\paragraph{Commitment and net‑zero ambition:} 
The correlation between commitment and quantitative target adoption is also weak and positive ($\rho=0.25$), suggesting that companies declaring commitments are somewhat more likely to disclose reduction or net‑zero targets, although the relationship is far from systematic.  
This highlights a persistent gap between strategic declarations and measurable climate objectives.

\paragraph{Specificity and net‑zero ambition:} 
Specificity and quantitative target adoption show the highest positive correlation among the pairs studied, but it remains weak ($\rho=0.34$).  
This indicates that detailed commitments are more frequently associated with the formalization of emission reduction or net‑zero objectives, but the link is still limited.\\
\noindent Although all correlations are statistically significant due to very low p‑values, their magnitude remains modest.  
This suggests that climate narrative tone, commitment expression, specificity, and quantitative target adoption represent distinct but partially connected dimensions of corporate climate communication.  
Risk‑oriented narratives are more often associated with explicit and detailed commitments, but translating these into measurable reduction or net‑zero goals remains an additional step that many firms have yet to take.


\subsubsection{Descriptive relations}

This subsection examines how climate commitment and the specificity of disclosures are associated with the tone of climate discourse (sentiment) and the adoption of quantitative climate targets (reduction and net-zero). 
The classification thresholds applied (40\% for commitment and specificity, 30\% for sentiment and target categories) provide a stricter, proportion-based definition of narrative attributes.\\

\noindent Overall, 86.5\% of firms communicate explicit climate commitments, and 77.6\% provide specific, verifiable disclosures (Table~\ref{tab:global-distribution}). 
From a sentiment perspective, 41.2\% of narratives adopt a risk framing, 32.7\% remain neutral, and only 5.2\% are opportunity-oriented. 
Only 0.8\%  exhibits both risk- and opportunity-oriented elements.\\

\noindent In terms of quantitative ambition, a strong majority (76.2\%) disclose net-zero targets, while 17.4\% lack any reduction objective and only 6.4\% pursue intermediate reduction pathways, either standalone (2.4\%) or in combination with net-zero goals (4.0\%).\\

\noindent \paragraph{Commitment versus sentiment, specificity and net-zero.}  
Firms without explicit commitments (13.5\% of the sample) tend to communicate differently: two-thirds adopt a neutral tone (66.1\%) and more than a quarter (27.7\%) adopt an opportunity framing, while risk- and risk-opportunity-oriented communication are rare (3.6\% and 2.7\%) (Table~\ref{tab:commitment-vs-sentiment-specificity-netzero}).  
In contrast, firms with explicit climate commitments are largely risk-oriented (59.1\%), 
with a significant proportion maintaining a neutral tone (37\%). 
Only a marginal 3.2\% adopt an opportunity framing, 
and just 0.7\% exhibit mixed risk-opportunity narratives.  
Most committed firms provide specific disclosures (83.1\%), whereas less than half of non-committed firms do so (42.9\%) (Table~\ref{tab:commitment-vs-sentiment-specificity-netzero}).  This result highlights an apparent paradox between commitment and specificity. Commitment captures whether firms explicitly communicate climate mitigation or adaptation commitments, whereas specificity differentiates between general statements and specific, verifiable climate actions (\cite{ISAB2018}). The finding that 42.9\% of non-committed firms nevertheless provide specific disclosures suggests that some firms describe concrete climate-related actions without framing them as formal commitments. This raises important questions about whether such disclosures represent credible action, symbolic communication, or sector-wide mimetic behaviors, thereby enriching the debate on how firms “talk their walk.”

\noindent In terms of target setting, 80\% of committed firms disclose net-zero objectives, 2.7\% adopt reduction-only targets, and 4.6\% combine both reduction and net-zero objectives. By contrast, 12.7\% report no quantitative climate target (Table~\ref{tab:commitment-vs-sentiment-specificity-netzero}).  
As for specificity, among non-committed firms the pattern is paradoxical too. A total of 51.8\% disclose net-zero objectives, 0.9\% adopt reduction-only targets, while only 47.3\% report no quantitative climate target, with no instance of combined reduction and net-zero objectives (Table~\ref{tab:commitment-vs-sentiment-specificity-netzero}). This surprising presence of ambitious net-zero targets among firms without explicit commitments mirrors the paradox observed in specificity, where firms without formal commitments nevertheless provide detailed disclosures. In line with our theoretical framework, such patterns suggest that target-setting, like disclosure specificity, does not always align with formal commitment. They may instead reflect symbolic communication, sector- or size-related drivers of disclosure maturity, or mimetic behaviors whereby non-committed firms mirror the practices of committed peers. This strengthens the argument for distinguishing between credible and symbolic climate narratives when assessing whether firms truly “talk their walk.”

\noindent \paragraph{Specificity versus target-setting:}  
A similar pattern emerges when comparing disclosure specificity and target-setting. Specific disclosures are strongly correlated with target formalization, with 83.4\% of specific firms adopting net-zero targets and only 9.2\% reporting no quantitative target (Table~\ref{tab:netzero-vs-specifcity}). 
Non‑specific firms still show 45.9\% without any quantitative target and 51.4\% adopting net‑zero objectives, 
an unexpected pattern similar to that observed among non‑committed firms.
\noindent \paragraph{Sentiment versus net-zero ambition:}  
The tone of climate discourse varies across target-setting categories (Table~\ref{tab:netzero-vs-sentiment}).  
Within each category, risk-oriented narratives dominate, reaching particularly high shares for reduction-only (95.0\%) 
and combined reduction–net-zero objectives (78.8\%).  
Opportunity-oriented communication is mainly observed among firms with no quantitative targets (22.9\%), 
suggesting that an opportunity framing does not necessarily translate into measurable climate action.  
Neutral narratives are common both among firms with no targets (37.5\%) and among those with net-zero objectives (44.1\%), 
indicating that a neutral tone coexists with a wide range of target-setting practices.
\\

\noindent 
These findings indicate a clear maturity gradient. 
firms with explicit and specific climate commitments tend to adopt risk-oriented narratives and are more likely to formalize ambitious climate targets, especially net-zero goals.  
Firms without commitments or specificity often communicate in a neutral or opportunity-oriented manner and are less likely to set quantitative targets.  
Nonetheless, the presence of firms with opportunity-oriented narratives but no targets, committed firms still lacking targets, and even some non‑committed or non‑specific firms that nevertheless adopt ambitious targets, highlights potential gaps and irregularities between strategic declarations and operational implementation.


\subsubsection{Cluster analysis of climate communication profiles}

A KMeans clustering analysis was conducted using four climate communication dimensions 
(sentiment, commitment, specificity, and net-zero ambition) together with emission exposure classes 
for Scope~1 (EI class) and Scope~2 (EJ class). 
Scope~3 was excluded from the clustering because of high missing data.  
Sentiment, commitment, specificity, and net-zero ambition were coded according to their semantic order, 
while Scope~1 and Scope~2 classes were encoded numerically (0 = lowest emission to 7 = highest emission).  

The centroids of the resulting clusters in the original variable space are reported in 
Table~\ref{tab:cluster_centroids_original}.  
The analysis produced ten distinct clusters, with company distribution shown in 
Table~\ref{tab:cluster_counts}. 
These insights are based on cluster centroids, which capture the average profile of companies within each cluster. 
This centroid-based approach reveals structural patterns in climate communication but does not capture firm‑specific details, 
and should therefore be interpreted as indicative rather than exhaustive.

\paragraph{High-emission clusters (3 and 7):}  

Cluster~3 consists of firms with very high direct and indirect emissions 
(Scope~1 centroid = 4.62, Scope~2 centroid = 3.26), 
high levels of commitment and specificity, and strong net‑zero ambition (2.70).  
Cluster~7 presents a broadly similar profile, with high Scope~2 exposure (6.20) and medium Scope~1 emissions (2.68), 
but slightly lower commitment (0.92), slightly lower specificity (0.92), 
and moderately lower net‑zero ambition (2.08).  
Despite these differences, both clusters exhibit comparable overall narratives, 
combining explicit climate engagement with risk‑oriented tones and a focus on long‑term objectives.  
This suggests that, on average, highly emitting companies tend to communicate in similar ways, 
differing mainly in the strength of their commitments and targets rather than in the nature of their discourse.

\paragraph{Low-emission clusters (4 and 5):} 
Clusters~4 (50 companies) and~5 (39 companies) are composed of companies with low emission levels but differing communication profiles. 
Cluster~4 includes firms that have no commitments, minimal specificity (0.08), no net-zero ambition, and a sentiment leaning toward opportunity (2.38). 
These firms communicate potential business benefits rather than risk management. 
Cluster~5, however, includes companies with formal commitments (1.00) but no specificity (0.00) and minimal net-zero ambition (0.10), representing firms that declare intent without operational maturity. 
These differences illustrate that even among low-emission firms, engagement levels vary from optimistic but disengaged narratives to formally committed yet shallow disclosure practices, 
which may indicate a potential risk of greenwashing when public commitments are not accompanied by verifiable actions or ambitious quantitative targets.

\paragraph{Mixed or mainstream clusters (0, 2, 6, 9):} 
Clusters~0, 2, 6, and~9 represent mainstream corporate climate profiles but differ in tone and disclosure maturity. 
Clusters~0 and~2 both show explicit commitments and highly specific disclosures (1.00 each), as well as high net-zero ambition (2.97 and 2.93, respectively), 
but their narrative orientation diverges: Cluster~2 is purely risk-oriented (0.00), while Cluster~0 adopts a more neutral tone (2.03). 
Clusters~6 and~9 also display contrasting patterns: Cluster~6 lacks explicit commitments (0.00) yet provides specific disclosures and strong net-zero ambition (2.95), 
suggesting target-driven behavior without formal narrative framing. 
Conversely, Cluster~9 combines explicit commitments and ambitious net-zero targets (2.99) but relies on generic disclosures (0.00), 
indicating that ambition is not fully supported by detailed reporting.

\paragraph{Anomalous cluster (1):} 
Cluster~1 is an outlier because it lacks commitments and specificity (both 0.00) but still shows relatively high net-zero ambition (2.47) and a sentiment leaning toward opportunity (2.29). 
This profile suggests that some companies signal ambitious climate objectives, likely under external stakeholder pressure, while lacking robust internal governance or detailed disclosure frameworks. 
Such a pattern may indicate early-stage climate strategy development, where long-term objectives are announced before operational plans are formalized, 
or, in some cases, opportunistic communication practices aimed at reputation enhancement rather than concrete action---a situation often referred to as potential \textit{greenwashing}.

\paragraph{Conservative cluster (8):} 
Cluster~8 is characterized by explicit commitments and specific disclosures (both 1.00) but very low net-zero ambition (0.22) and a moderately risk-oriented tone (0.62). 
This reflects a communication style that emphasizes compliance and formalism rather than ambitious strategic goals.


\subsubsection{Integrated insights from correlation, descriptive, and cluster analyses}

The correlation analysis highlights statistically significant but weak associations between narrative tone, disclosure specificity, and target ambition, 
indicating that these elements of climate communication are related but only partially dependent. 

Descriptive cross‑tabulations show how these relationships manifest in practice, revealing a maturity gradient: 
explicit and specific commitments are frequently associated with risk‑oriented narratives and a higher prevalence of net‑zero targets, 
whereas firms adopting neutral or opportunity‑oriented tones more often lack quantified objectives. 
These descriptive findings also expose irregularities that correlations alone cannot fully capture, 
such as firms without commitments adopting ambitious targets or firms with explicit commitments still lacking measurable objectives.

Cluster analysis further confirms that companies often converge around a limited number of disclosure styles, 
despite differences in emissions or organizational characteristics. 
This similarity of communication patterns among otherwise different firms suggests potential mimicry behavior, 
where companies replicate peers’ disclosure styles without necessarily aligning them with substantive climate strategies \citep{huang2025imitation, AERTS2006299}. 
Overall, these integrated results highlight the coexistence of voluntary and legitimacy-driven motives and point to structural gaps 
between narrative framing, commitment expression, and operational climate objectives.


\subsection{Firm-Level climate communication by size and sector}

\paragraph{Discourse tone:}  
 
The overall tone of climate-related corporate narratives shows a predominance of risk-oriented communication (41.2\%), 
followed by neutral narratives (32.7\%), while only 5.2\% adopt an explicitly opportunity-oriented perspective 
and 0.8\% exhibit mixed risk–opportunity framing.  
This distribution indicates that corporate climate discourse remains primarily focused on risk management and compliance rather than portraying climate change as a strategic opportunity.

\textit{Firm size:}  
Across market capitalization classes, risk-oriented sentiment becomes more prevalent as firm size increases.  
The largest firms (Cap\_7 and Cap\_8) display the highest shares of risk communication (61.0\% and 76.2\%, respectively), while their neutral tone decreases significantly to 32.2\% and 23.8\%.  
Conversely, smaller firms (Cap\_1–Cap\_4) maintain a more balanced (\textit{risk- neutral}) profile, with neutral sentiment reaching up to 49.7\% (Cap\_2).  
Opportunity-oriented narratives are relatively more frequent among smaller firms (Cap\_1 to Cap\_3) where they reach up to about 9\%, 
but their prevalence decreases steadily with firm size, becoming almost insignificant among the largest capitalization classes.  
Risk–opportunity mixed narratives, by contrast, remain marginal across all classes (0–1.6\%).  
This pattern indicates that smaller firms are relatively more inclined to frame climate action as a potential source of growth, 
whereas such a perspective is almost entirely absent among larger firms.\\

A similar pattern emerges when firm size is measured by number of employees.  
Risk-oriented discourse increases steadily with workforce size, rising from 34.1\% in the smallest employee class (Emp\_01) 
to 78.6\% in the largest (Emp\_08), while neutral sentiment declines from 51.2\% to just 7.1\%.  
Opportunity-oriented narratives are relatively more common among smaller firms (12.5--14.6\% for Emp\_01 and Emp\_02) 
but become marginal or virtually absent among largest firms.  
This distribution suggests that larger organizations, likely subject to greater regulatory scrutiny and stakeholder pressure, 
tend to adopt a more cautious and risk‑focused tone rather than framing climate change as a potential source of growth.\\

\noindent \textit{Emission intensity - scopes 1,2 and 3 (EI, EJ, EK):}  
Across all three scope classifications (EI, EJ, EK) a similar trend emerges.  
Firms with the lowest emission intensity (classes C1–C2) display a nearly balanced narrative 
with risk and neutral tones at comparable levels (approximately 43–46\%).  
As emission intensity rises this balance shifts progressively toward risk-oriented communication, 
reaching 56.9–60.8\% in medium-intensity classes (C3–C5) and peaking at very high levels for the most emission‑intensive firms 
(77.1\% and 80.0\% in EI‑C7 and EI‑C8, 90.5\% in EJ‑C6, 75.0\% in EK‑C7 and 73.2\% in EK‑C8).  
Neutral tone follows the opposite trajectory, declining steadily as emission intensity increases.  
Opportunity-oriented narratives represent around 10\% of discourse among the lowest-intensity firms for each scope, 
a non‑negligible share (like that observed in the Finance sector and the smallest firm sizes), 
but they become marginal or nearly absent among medium and higher-emission firms.  
This common progression from an almost balanced risk–neutral profile in low emitters 
to a strong dominance of risk narratives in high emitters 
shows that greater climate exposure systematically drives firms toward defensive and risk‑focused communication.  \\

\textit{Sectoral differences:}  
The Energy sector surprisingly shows one of the highest neutral shares (61.0\%) despite its high climate exposure, 
suggesting highly descriptive rather than risk-driven communication.  
Communication, Suppliers, and Industrial sectors present a strong risk orientation (66.7\%, 66.7\%, and 60.3\%, respectively), 
reflecting exposure to technological disruption and value‑chain dependencies.  
By contrast, Finance, Real Estate, and Health Care exhibit relatively low levels of risk-oriented communication 
(36.5\%, 36.7\%, and 38.1\%, respectively) and higher shares of neutral tone 
(43.2\% for Finance, 58.3\% for Real Estate, and 49.2\% for Health Care).  
Finance stands out with the highest opportunity-oriented share (17.6\%), 
indicating that financial firms are more likely to frame climate change as a potential growth opportunity.  
For all other sectors, risk-oriented narratives dominate clearly (above 55\%), 
with neutral tones limited to less than 42\% and opportunity framing remaining marginal.  
Overall, while most sectors adopt risk-oriented narratives, 
Finance, Real Estate, and Health Care rely more on descriptive or opportunity‑focused communication.

\paragraph{Commitment expression:}  
A striking feature of firm-level climate communication is the high prevalence of explicit climate commitments: overall, 86.5\% of firms disclose at least one climate-related commitment, while only 13.5\% have no explicit engagement (Table~\ref{tab:global-distribution}).  
However, the depth and framing of these commitments vary substantially across firm size, emission intensity, and sector.  

\textit{Firm size:}  
Commitment levels increase with market capitalization, suggesting that larger firms may experience stronger external pressure from regulators, institutional investors, and civil society.  
In the smallest capitalization class (Cap\_1), 80.6\% of firms disclose commitments, rising steadily to 100\% in the largest class (Cap\_8).  
This gradient reflects the resources available to large firms to formalize climate strategies and the higher expectations they face due to their systemic importance.  
A similar pattern emerges when firm size is measured by workforce: commitment rates are lowest in Emp\_02 (71.2\%) but reach 94.3\% in Emp\_06 and 85.7\% in Emp\_08.  
These findings imply that both financial and organizational capacity influence the likelihood of formal climate commitments, with smaller firms perhaps constrained by resources, expertise, or governance maturity.  

\textit{Emission intensity:}  
Commitment rates increase consistently with emission exposure across all three scope classifications (EI, EJ, and EK).  
For the lowest-intensity classes (C1–C2) commitment levels remain relatively modest 
(82.3\% for EI and comparable levels for EJ and EK), 
but they rise steadily with emission intensity, exceeding 90\% from class C3 upward 
and reaching full coverage in some of the highest-emission classes 
(100\% in EJ‑C6 and EJ‑C7, and 97.6\% in EK‑C8).  
This common pattern indicates that firms with higher direct, indirect, or value‑chain emissions 
are more likely to formalize climate commitments, reflecting stronger regulatory pressure, 
increased reputational risk, and heightened stakeholder expectations linked to their greater climate impact.

\textit{Sectoral patterns:}  
Sector differences are also visible.  
Materials (95.7\%), Industry (90.4\%), and Cyclical Consumption (90.3\%) lead in commitment rates, suggesting that sectors with tangible production footprints and visible supply chains are particularly responsive to stakeholder and regulatory pressure.  
In contrast, Energy (74.6\%) and Finance (77.0\%) lag, which may reflect a strategic reluctance to adopt formal climate positions in highly regulated (Energy) or indirect-impact (Finance) industries.  
Communication and Real Estate maintain relatively high commitment rates (86.7\% and 88.3\%), illustrating that even service-oriented sectors increasingly recognize the need to publicly address climate risks.  \\

\noindent Taken together, these results highlight that explicit climate commitments are widespread and correlated with firm size, emission intensity, and sector exposure.  
However, the existence of non-committed firms even in high-intensity or high-visibility sectors indicates that commitment communication is not fully determined by objective exposure, suggesting differences in governance maturity, strategic orientation, or managerial risk perception.  

\paragraph{Commitment specificity:}  
While commitment expression is widespread, the quality and precision of these commitments are also high and closely aligned with overall commitment levels.  
Overall, 77.6\% of firms provide specific commitments (86.5\% for commitment), while 22.4\% remain vague or generic (Table~\ref{tab:global-distribution}).  
This proximity between commitment and specificity rates indicates that most firms not only declare climate intentions but also articulate them in measurable terms, 
signaling a willingness to be held accountable for their actions. \\

\noindent \textit{Firm size:}  
Specificity is clearly linked to firm size.  
Among the smallest capitalization firms (Cap\_1), only 68.5\% provide specific commitments, compared to 95.2\% in the largest class (Cap\_8).  
Similarly, specificity rises from 73.2\% in Emp\_01 to 92.9\% in Emp\_08.  
These patterns suggest that larger firms not only commit more often but also tend to articulate their commitments in measurable terms, likely due to stronger governance structures, greater reporting capabilities, and heightened public scrutiny.  \\

\noindent \textit{Emission intensity:}  
Specificity is also associated with emission profiles.  
In Scope~EI, specificity rises from 66.3\% in C1 to 100\% in C8, reflecting stronger disclosure expectations for high emitters.  
A similar trend appears in Scope~EJ and EK, where the share of specific commitments exceeds 90\% in the top classes.  
This indicates that firms facing greater climate risks—either operational (Scope~EI), indirect (Scope~EJ), or value-chain related (Scope~EK)—are more likely to provide measurable and auditable commitments, possibly due to stakeholder pressure or anticipation of regulation.  \\

\noindent \textit{Sectoral variation:}  
Specificity rates vary substantially across sectors.  
Suppliers (97.8\%) and Materials (88.4\%) lead, likely due to pressure from value‑chain partners and carbon‑intensive processes, 
whereas Finance (59.5\%) and Health Care (68.3\%) lag, reflecting more indirect climate exposure and lower perceived necessity to disclose operational targets.  
Energy, despite high climate exposure, only reaches 69.5\% specificity, suggesting that some firms in this sector may prefer general statements, 
possibly to maintain strategic flexibility amid regulatory uncertainty.  
These patterns broadly mirror commitment levels across sectors but reveal that, in Finance, Health Care, and Energy, 
the quality of disclosure (specificity) is lower relative to their commitment rates.  
\\

\noindent These results indicate that commitment specificity follows similar patterns to overall commitment: larger, high-emission, and industrial firms tend to provide more detailed disclosures.  
However, the existence of vague commitments in some highly exposed sectors suggests differences in reporting culture and corporate climate strategies, potentially linked to reputational risk management or the desire to retain operational discretion.  

\paragraph{Target setting:}  
Target adoption provides another dimension of climate strategy maturity.  
Overall, 76.2\% of firms adopt net-zero targets, 17.4\% report no quantitative targets, and only 2.4\% adopt only reduction targets, and 4.0\% combine reduction and net-zero goals (Table~\ref{tab:global-distribution}).  
This high prevalence of net-zero pledges highlights the growing importance of long-term climate neutrality in corporate strategies.  

\textit{Firm size:}  
The share of firms without any quantitative climate target tend gloabally to decreases slightly with both market capitalization and workforce size 
(maximum 27.4\% in Cap\_1 and minimum 11.9\% in Cap\_7, 
while for workforce classes the maximum is 24.4\% in Emp\_01 and the minimum 7.1\% in Emp\_08). 
This indicates that the proportion of firms adopting at least one climate objective increases modestly as firms grow.  
Net-zero targets show different concentration patterns. While net-zero adoption gradually rises with workforce size 
(from 70.7\% in Emp\_01 to 78.6–85.1\% in Emp\_06–Emp\_08), 
market capitalization exhibits a slight concentration in medium classes.  
These patterns suggest that firms with larger workforces are more likely to adopt net‑zero targets, 
indicating both a stronger orientation toward long‑term climate objectives and a greater capacity—or external pressure—to formalize such strategies.  
Mid‑capitalization firms also show a relatively strong concentration of net‑zero pledges despite their smaller absolute size, 
suggesting that factors beyond financial scale, such as stakeholder expectations or sectoral positioning, may drive climate ambition.

\textit{Emission intensity:}  
Target adoption patterns vary with emission exposure.  
For both Scope~1 and Scope~2, the share of firms without any reduction target follows a U‑shaped pattern, implying an inverted U‑shape for firms adopting at least one target.  
Net‑zero commitments, which represent the largest share of targets, also display an inverted U‑shape, with a notable concentration in intermediate emission classes.  
For Scope~3, by contrast, the share of firms without targets decreases slightly with emission intensity, indicating a modest increase in the proportion of firms adopting at least one target.  
Net‑zero commitments in Scope~3 exhibit a mild inverted U‑shape but are more prevalent in the higher‑emission classes, suggesting that companies facing complex value‑chain emissions are progressively more inclined to commit to long‑term climate neutrality.  

\textit{Sectoral differences:}  
Net‑zero adoption is particularly strong in Energy (86.4\%), Basic Materials (87.0\%), and Real Estate (81.7\%), sectors generally subject to intense regulatory and investor scrutiny.  
Finance (59.5\%) and Communication (63.3\%) show lower uptake, consistent with their more indirect emission profiles and weaker operational incentives to set such targets.  
Reduction‑only targets remain rare across all sectors (0 - 4.4\%), and most companies opting for quantified goals move directly toward long‑term net‑zero commitments, often supplemented by interim reduction (reduction-Netzero) milestones.\\

\noindent \textbf{Summary:}  
Climate target setting has become mainstream, with most firms committing to long‑term net‑zero objectives rather than incremental reductions.  
Workforce‑intensive and mid‑capitalization firms show higher engagement, and companies with complex value‑chain emissions are increasingly committing to net‑zero.  
Sectoral differences persist, with Energy, Basic Materials, and Real Estate leading adoption, while Finance and Communication remain less engaged.  
Overall, climate objectives are no longer limited to a small group of leaders but represent a broad strategic shift across diverse industries and firm profiles.



\subsection{Cross-dimensional insights and implications}

\paragraph{A consistent baseline with emerging differentiation:} 
A relatively uniform regulatory and market environment provides a common baseline for climate discourse across US-listed firms, as reflected in the high prevalence of climate commitments (70.3\%) and the dominant neutral or risk-oriented tone. 
However, differences emerge when firm size, sectoral exposure, and emission intensity are considered. 
Market capitalization, rather than employee count, is a more reliable predictor of disclosure depth, with larger firms tending to adopt more risk-aware narratives, higher specificity (up to 38.9\% in the largest capitalization class), and a greater prevalence of net-zero commitments. 
Similarly, emission class adds another layer of differentiation: Scope~1 and Scope~2 companies display clearer upward trends in target adoption, while Scope~3 companies show more heterogeneous behaviors.

\paragraph{Communication patterns and structural capacity:} 
The relationship between narrative tone, commitment specificity, and target setting reveals distinct strategic approaches. 
Firms with explicit commitments communicate more frequently in risk-oriented terms (57.9\%) and are more likely to specify their pledges (28.4\% vs 4.6\% for non-committed firms). 
These firms are also more likely to formalize quantitative objectives, including net-zero targets (12.1\% vs 4.6\% for non-committed firms). 
Nevertheless, gaps remain: 36\% of committed firms still have no explicit targets, and 35.5\% of firms with specific commitments lack any target, highlighting a disconnect between strategic declarations and operational implementation. 
Conversely, 45.8\% of non-committed firms disclose reduction targets, indicating that some companies adopt climate measures for compliance or reputational reasons without framing them as part of a formal climate commitment strategy.

\paragraph{Scope-based maturity gradients:} 
Emission intensity plays a significant role in shaping disclosure practices. 
For Scope~1 and Scope~2 emissions, the proportion of firms without climate targets decreases substantially (from 46.4\% to 25\% for Scope~1, and from 45.1\% to 26.6\% for Scope~2), while net-zero commitments increase notably in the highest-emitting classes (C7 and C8). 
In contrast, for Scope~3 emissions, no clear pattern emerges in the adoption of reduction or combined targets, indicating that value chain emissions remain complex and inconsistently addressed across firms.
However, net-zero ambitions increase across all three scopes, confirming that higher emitters tend to set more ambitious long-term goals.

\paragraph{Policy and analytical relevance:} 
These findings imply that a one-size-fits-all approach to ESG policy or evaluation may overlook key nuances. 
While most firms comply with baseline expectations (neutral tone, broad commitments, some reduction targets), only a subset of firms—typically larger and higher-emitting ones—translate these commitments into specific pledges and ambitious targets. 
This suggests that ESG evaluation frameworks should consider not only whether firms commit but also how they communicate (tone, specificity) and what they operationalize (target type). 
For policymakers, stronger incentives or differentiated requirements may be needed to convert broad statements into actionable, measurable objectives, particularly among mid-sized and Scope~3-dominant firms.

\paragraph{Scholarly contribution:} 
This study demonstrates the analytical power of large language models (LLMs) in mapping corporate climate discourse at scale. 
It identifies consistent baseline behaviors alongside structural differentiation driven by firm size, emission intensity, and sectoral exposure. 
By linking narrative tone, commitment specificity, and target setting, this research highlights disclosure asymmetries even in regulated environments and shows how narrative analysis can capture ESG integration maturity beyond traditional quantitative metrics.


\section{Conclusion and policy implications}

\subsection{Cross-sectional insights}

This study provides evidence that differences in corporate climate communication narratives—captured through sentiment, commitment, specificity, and target ambition—exist across firm size, sector, and emission profiles, but these differences are often modest.  
The weak overall variation suggests two main explanations: (i) mimicking behavior, where firms replicate peers’ disclosure styles regardless of their own climate maturity, and (ii) limited disclosure sophistication, where firms adopt symbolic or generic communication strategies rather than fully integrated climate transition narratives.  
In some cases, these patterns are consistent with greenwashing, where ambitious claims are made without operational depth or alignment with measurable objectives.

\subsection{Hypothesis validation}

H1 is supported.  
Highly emitting firms (e.g., Clusters~3 and~7) tend to express explicit commitments and relatively specific disclosures but show comparatively lower adoption of reduction or net-zero objectives.  
Descriptive results confirm that firms with the highest emissions are generally more likely to communicate commitments and detailed actions, but these commitments often fall short of ambitious transition objectives compared to low and medium emitting firms.

H2 is not supported.  
Contrary to expectations, the lowest-emission firms show the lowest levels of commitments, specificity, and ambition regarding reduction or net-zero targets.  
Clustering results reveal a strong disparity among low-emission firms,  commitments  and specificity.  
This finding indicates that lower emissions do not necessarily translate into more mature or credible climate disclosure practices.

H3 is partially supported.  
Inconsistencies between narrative tone (e.g., optimistic or opportunity-oriented framing) and the presence of concrete commitments or targets are observed, but they are relatively marginal in our sample.  
Such inconsistencies, although not widespread, indicate areas where narrative positioning and operational objectives are not fully aligned.

H4 and H5 are validated, although differences by firm size and sector remain relatively small.  
Larger firms tend to adopt more explicit commitments and higher-quality disclosures, while smaller firms exhibit more heterogeneous communication practices.  
Sectoral variation is visible but limited in magnitude.  
For instance, the Energy sector, despite high climate exposure, shows a high share of neutral narratives (61\%), whereas Finance has the highest opportunity-oriented share (18\%), framing climate change more as a business opportunity.  
Sectors with tangible production footprints, such as Materials and Industry, lead in commitment rates (above 90\%), while Finance and Energy lag (below 78\%).  
Specificity is highest in supplier-dependent sectors but remains lower in Finance and Energy, and net‑zero targets are most common in Energy and Basic Materials (over 85\%) compared to Finance (60\%).  
Overall, while sector and size affect disclosure maturity, the differences are relatively modest, suggesting peer influence and convergence in reporting practices.

H6 is supported, at least partially.  
The limited differences observed across emissions, firm size, and sector, combined with the strong similarity of narrative types across clusters, suggest that institutional pressures and peer benchmarking drive convergence in disclosure practices, indicating potential mimicry effects in climate communication.

\subsection{Policy implications}

The convergence of climate narratives, limited differentiation by firm characteristics, and presence of symbolic or inconsistent disclosure practices suggest that voluntary reporting alone may be insufficient to ensure credible and decision-useful climate information.  
Regulators and standard-setters should focus on strengthening requirements for specificity and target linkage, ensuring that ambitious announcements are supported by verifiable interim goals and operational plans.  
Sector-specific disclosure guidelines and third-party assurance can help address mimicry by linking communication explicitly to performance metrics.  
Improving digital taxonomies and encouraging alignment with frameworks such as TCFD and SBTi could further enhance comparability and credibility, limiting opportunities for symbolic disclosure and greenwashing.

\subsection{Limitations and future research}

This analysis relies on a cross-sectional dataset and does not capture how disclosure strategies evolve over time.  
We believe that future time-series studies should consider longer observation windows (e.g., four years or more), since corporate narratives often take substantial time to change.  
Thresholds used for aggregating paragraph-level classifications, although designed for robustness, may introduce measurement biases.  
During this research, we also explored estimating emission levels directly from LLM-derived disclosure features using various approaches, including fuzzy logic.  
However, the high level of narrative similarity and mimicry identified in this study suggests that such indirect estimation is not feasible without additional performance-based data.  
Future work could extend this research by integrating longitudinal analysis of disclosure maturity, comparing narrative-based classifications with actual climate performance, and applying richer embedding-based text analysis to capture disclosure nuance beyond predefined classification labels.


\section{Appendix}


\begin{table}[htbp]
\centering
\caption{Nombre de NaN par colonne}
\begin{tabular}{lr}
\hline
\textbf{Colonne} & \textbf{Nombre de NaN} \\ \hline
EI                                & 0   \\
EJ                                & 15  \\
EK                                & 355 \\
Employé                           & 0   \\
Capitalisation boursière\_Mrd     & 0   \\
sentiment\_global                 & 0   \\
commitment\_global                & 0   \\
specificity\_global               & 0   \\
netzero\_global                   & 0   \\ \hline
\end{tabular}
\label{tab:nan_counts}
\end{table}

\vspace{0.5cm}

\begin{table}[htbp]
\centering
\caption{Statistiques descriptives des variables numériques (1/2)}
\begin{tabular}{lrrrrr}
\hline
\textbf{Variable} & \textbf{Count} & \textbf{Mean} & \textbf{Std} & \textbf{Min} & \textbf{Max} \\ \hline
EI                              & 828 & 1,627,959  & 6,319,275  & 0        & 84,875,100  \\
EJ                              & 813 & 368,106    & 1,167,455  & 0        & 17,741,000  \\
EK                              & 473 & 19,508,980 & 78,710,990 & 0        & 1,169,970,000 \\
Employé                         & 828 & 27,300.3   & 98,142.3   & 2        & 2,300,000   \\
Cap\_Mrd   & 828 & 48.20      & 214.93     & 0.00186  & 2,960.0     \\ \hline
\end{tabular}
\label{tab:stats_num_part1}
\end{table}

\begin{table}[htbp]
\centering
\caption{Statistiques descriptives des variables numériques (2/2)}
\begin{tabular}{lrrr}
\hline
\textbf{Variable} & \textbf{25\%} & \textbf{50\%} & \textbf{75\%} \\ \hline
EI                              & 3,710     & 27,444   & 280,976  \\
EJ                              & 9,483     & 42,677   & 196,300  \\
EK                              & 108,086   & 1,116,520& 7,769,050 \\
Employé                         & 2,081     & 7,138    & 20,000   \\
Cap\_Mrd   & 2.2625    & 7.240    & 26.0625  \\ \hline
\end{tabular}
\label{tab:stats_num_part2}
\end{table}

\vspace{0.5cm}

\begin{table}[htbp]
\centering
\caption{Statistiques descriptives des variables catégorielles}
\begin{tabular}{lrrrr}
\hline
\textbf{Variable} & \textbf{Count} & \textbf{Unique} & \textbf{Top} & \textbf{Freq} \\ \hline
sentiment\_global   & 828 & 4 & risk        & 427 \\
commitment\_global  & 828 & 2 & commitment  & 716 \\
specificity\_global & 828 & 2 & specific    & 643 \\
netzero\_global     & 828 & 4 & netzero     & 631 \\ \hline
\end{tabular}
\label{tab:stats_cat}
\end{table}



\begin{table}[htbp]
\centering
\caption{Category encoding (semantic order) for categorical LLM labels.
Thresholds used: commitment $>$ 40\%, specificity $>$ 40\%, risk $>$ 30\%, opportunity $>$ 30\%, net-zero $>$ 30\%, reduction $>$ 30\%.}
\begin{tabular}{ll}
\hline
\textbf{Variable} & \textbf{Encoding} \\ \hline
sentiment\_global & 0: risk,\; 1: risk\_opportunity,\; 2: neutral,\; 3: opportunity \\ 
commitment\_global & 0: no\_commitment,\; 1: commitment \\ 
specificity\_global & 0: general,\; 1: specific \\ 
netzero\_global & 0: no\_reduction,\; 1: reduction,\; 2: reduction\_netzero,\; 3: netzero \\ \hline
\end{tabular}
\label{tab:semantic-encoding}
\end{table}

\begin{table}[htbp]
\centering
\caption{Spearman correlation p-values (semantic order, compact form, small $\approx$ 0 replaced by 0.00). Thresholds used: commitment $>$ 40\%, specificity $>$ 40\%, risk $>$ 30\%, opportunity $>$ 30\%, net-zero $>$ 30\%, reduction $>$ 30\%.}
\begin{tabular}{lrrrr}
\hline
 & \textbf{Sentiment} & \textbf{Commitment} & \textbf{Specificity} & \textbf{Netzero} \\ \hline
Sentiment & 0.00 & 0.00 & 0.00 & 0.007 \\
Commitment & 0.00 & 0.00 & 0.00 & 0.00 \\
Specificity & 0.00 & 0.00 & 0.00 & 0.00 \\
Netzero   & 0.007 & 0.00 & 0.00 & 0.00 \\ \hline
\end{tabular}
\label{tab:spearman-pvalues-compact}
\end{table}

\begin{table}[H]
\centering
\caption{Overall distribution of sentiment, commitment, specificity and net-zero/reduction classifications. 
Thresholds used: commitment $>$ 40\%, specificity $>$ 40\%, risk $>$ 30\%, opportunity $>$ 30\%, net-zero $>$ 30\%, reduction $>$ 30\%.}
\begin{tabular}{lrr}
\hline
\textbf{Sentiment} & \textbf{Count} & \textbf{\%} \\ \hline
Neutral & 339 & 32.7\% \\
Opportunity & 54 & 5.2\% \\
Risk & 427 & 41.2\% \\
Risk\_Opportunity & 8 & 0.8\% \\ \hline
\textbf{Commitment} & \textbf{Count} & \textbf{\%} \\ \hline
No Commitment & 112 & 13.5\% \\ 
Commitment & 716 & 86.5\% \\ \hline
\textbf{Specificity} & \textbf{Count} & \textbf{\%} \\ \hline
General & 185 & 22.4\% \\
Specific & 643 & 77.6\% \\ \hline
\textbf{Net-Zero/Reduction} & \textbf{Count} & \textbf{\%} \\ \hline
Net Zero & 631 & 76.2\% \\
No Reduction & 144 & 17.4\% \\
Reduction & 20 & 2.4\% \\
Reduction\_NetZero & 33 & 4.0\% \\ \hline
\end{tabular}
\label{tab:global-distribution}
\end{table}

\begin{table}[H]
\centering
\caption{Distribution of sentiment, specificity and net-zero/reduction by commitment status.
Thresholds used: commitment $>$ 40\%, specificity $>$ 40\%, risk $>$ 30\%, opportunity $>$ 30\%, net-zero $>$ 30\%, reduction $>$ 30\%.}
\begin{tabular}{lrr}
\hline
\textbf{Sentiment (Committed)} & \textbf{Count} & \textbf{\%} \\ \hline
Neutral            & 265 & 37.0\% \\ 
Opportunity        & 23  & 3.2\%  \\ 
Risk               & 423 & 59.1\% \\ 
Risk\_Opportunity  & 5   & 0.7\%  \\ \hline
\textbf{Sentiment (Non-committed)} & \textbf{Count} & \textbf{\%} \\ \hline
Neutral            & 74  & 66.1\% \\ 
Opportunity        & 31  & 27.7\% \\ 
Risk               & 4   & 3.6\%  \\ 
Risk\_Opportunity  & 3   & 2.7\%  \\ \hline
\textbf{Specificity (Committed)} & \textbf{Count} & \textbf{\%} \\ \hline
General            & 121 & 16.9\% \\ 
Specific           & 595 & 83.1\% \\ \hline
\textbf{Specificity (Non-committed)} & \textbf{Count} & \textbf{\%} \\ \hline
General            & 64  & 57.1\% \\ 
Specific           & 48  & 42.9\% \\ \hline
\textbf{Net-zero/Reduction (Committed)} & \textbf{Count} & \textbf{\%} \\ \hline
Net Zero           & 573 & 80.0\% \\ 
No Reduction       & 91  & 12.7\% \\ 
Reduction          & 19  & 2.7\%  \\ 
Reduction\_NetZero & 33  & 4.6\%  \\ \hline
\textbf{Net-zero/Reduction (Non-committed)} & \textbf{Count} & \textbf{\%} \\ \hline
Net Zero           & 58  & 51.8\% \\ 
No Reduction       & 53  & 47.3\% \\ 
Reduction          & 1   & 0.9\%  \\ \hline
\end{tabular}
\label{tab:commitment-vs-sentiment-specificity-netzero}
\end{table}


\begin{table}[H]
\centering
\caption{Distribution of sentiment by specificity status.
Thresholds used: specificity $>$ 40\%, risk $>$ 30\%, opportunity $>$ 30\%.}
\begin{tabular}{lrr}
\hline
\textbf{Sentiment (General)} & \textbf{Count} & \textbf{\%} \\ \hline
Neutral            & 90  & 48.6\% \\ 
Opportunity        & 42  & 22.7\% \\ 
Risk               & 49  & 26.5\% \\ 
Risk\_Opportunity  & 4   & 2.2\%  \\ \hline
\textbf{Sentiment (Specific)} & \textbf{Count} & \textbf{\%} \\ \hline
Neutral            & 249 & 38.7\% \\ 
Opportunity        & 12  & 1.9\%  \\ 
Risk               & 378 & 58.8\% \\ 
Risk\_Opportunity  & 4   & 0.6\%  \\ \hline
\end{tabular}
\label{tab:specificity-vs-sentiment}
\end{table}

\begin{table}[H]
\centering
\caption{Distribution of net-zero/reduction by specificity status.
Thresholds used: commitment $>$ 40\%, specificity $>$ 40\%, risk $>$ 30\%, opportunity $>$ 30\%, net-zero $>$ 30\%, reduction $>$ 30\%.}
\begin{tabular}{lrr}
\hline
\textbf{Net-Zero/Reduction (Specific)} & \textbf{Count} & \textbf{\%} \\ \hline
Net Zero & 536 & 83.4\% \\
No Reduction & 59 & 9.2\% \\
Reduction & 16 & 2.5\% \\
Reduction\_NetZero & 32 & 5.0\% \\ \hline
\textbf{Net-Zero/Reduction (Non-specific)} & \textbf{Count} & \textbf{\%} \\ \hline
Net Zero & 95 & 51.4\% \\
No Reduction & 85 & 45.9\% \\
Reduction & 4 & 2.2\% \\
Reduction\_NetZero & 1 & 0.5\% \\ \hline
\end{tabular}
\label{tab:netzero-vs-specifcity}
\end{table}

\begin{table}[htbp]
\centering
\caption{Distribution of NetZero vs Sentiment}
\begin{tabular}{lrrrr}
\hline
\textbf{NetZero Category} & \textbf{Risk} & \textbf{Risk\_Opportunity} & \textbf{Neutral} & \textbf{Opportunity} \\ \hline
No Reduction        & 52 (36.1\%) & 5 (3.5\%) & 54 (37.5\%) & 33 (22.9\%) \\
Reduction           & 19 (95.0\%) & 0 (0.0\%) &  1 (5.0\%)  &  0 (0.0\%) \\
Reduction\_Netzero  & 26 (78.8\%) & 1 (3.0\%) &  6 (18.2\%) &  0 (0.0\%) \\
Netzero             & 330 (52.3\%) & 2 (0.3\%) & 278 (44.1\%) & 21 (3.3\%) \\ \hline
\end{tabular}
\label{tab:netzero-vs-sentiment}
\end{table}

\begin{table}[htbp]
\centering
\caption{Distribution by Market Capitalization Classes}
\begin{tabular}{lrrrr}
\hline
\textbf{Sentiment} & Risk & Risk\_Opportunity & Neutral & Opportunity \\ \hline
Cap\_1 <1    & 59 (47.6\%) & 2 (1.6\%) & 51 (41.1\%) & 12 (9.7\%) \\
Cap\_2 1–4   & 83 (44.9\%) & 2 (1.1\%) & 92 (49.7\%) & 8 (4.3\%) \\
Cap\_3 4–10  & 83 (52.5\%) & 1 (0.6\%) & 59 (37.3\%) & 15 (9.5\%) \\
Cap\_4 10–20 & 58 (50.0\%) & 1 (0.9\%) & 49 (42.2\%) & 8 (6.9\%) \\
Cap\_5 20–40 & 51 (56.0\%) & 2 (2.2\%) & 35 (38.5\%) & 3 (3.3\%) \\
Cap\_6 40–80 & 41 (55.4\%) & 0 (0.0\%) & 29 (39.2\%) & 4 (5.4\%) \\
Cap\_7 80–240& 36 (61.0\%) & 0 (0.0\%) & 19 (32.2\%) & 4 (6.8\%) \\
Cap\_8 >240  & 16 (76.2\%) & 0 (0.0\%) & 5 (23.8\%)  & 0 (0.0\%) \\ \hline
\end{tabular}

\vspace{0.5cm}

\begin{tabular}{lrr}
\hline
\textbf{Commitment} & No Commitment & Commitment \\ \hline
Cap\_1 <1    & 24 (19.4\%) & 100 (80.6\%) \\
Cap\_2 1–4   & 30 (16.2\%) & 155 (83.8\%) \\
Cap\_3 4–10  & 18 (11.4\%) & 140 (88.6\%) \\
Cap\_4 10–20 & 16 (13.8\%) & 100 (86.2\%) \\
Cap\_5 20–40 & 10 (11.0\%) & 81 (89.0\%) \\
Cap\_6 40–80 & 6 (8.1\%)   & 68 (91.9\%) \\
Cap\_7 80–240& 8 (13.6\%)  & 51 (86.4\%) \\
Cap\_8 >240  & 0 (0.0\%)   & 21 (100.0\%) \\ \hline
\end{tabular}

\vspace{0.5cm}

\begin{tabular}{lrr}
\hline
\textbf{Specificity} & General & Specific \\ \hline
Cap\_1 <1    & 39 (31.5\%) & 85 (68.5\%) \\
Cap\_2 1–4   & 38 (20.5\%) & 147 (79.5\%) \\
Cap\_3 4–10  & 42 (26.6\%) & 116 (73.4\%) \\
Cap\_4 10–20 & 22 (19.0\%) & 94 (81.0\%) \\
Cap\_5 20–40 & 16 (17.6\%) & 75 (82.4\%) \\
Cap\_6 40–80 & 18 (24.3\%) & 56 (75.7\%) \\
Cap\_7 80–240& 9 (15.3\%)  & 50 (84.7\%) \\
Cap\_8 >240  & 1 (4.8\%)   & 20 (95.2\%) \\ \hline
\end{tabular}

\vspace{0.5cm}

\begin{tabular}{lrrrr}
\hline
\textbf{Net-Zero} & No Reduction & Reduction & Reduction\_Netzero & Netzero \\ \hline
Cap\_1 <1    & 34 (27.4\%) & 4 (3.2\%) & 2 (1.6\%) & 84 (67.7\%) \\
Cap\_2 1–4   & 28 (15.1\%) & 2 (1.1\%) & 4 (2.2\%) & 151 (81.6\%) \\
Cap\_3 4–10  & 32 (20.3\%) & 5 (3.2\%) & 7 (4.4\%) & 114 (72.2\%) \\
Cap\_4 10–20 & 14 (12.1\%) & 1 (0.9\%) & 5 (4.3\%) & 96 (82.8\%) \\
Cap\_5 20–40 & 15 (16.5\%) & 1 (1.1\%) & 2 (2.2\%) & 73 (80.2\%) \\
Cap\_6 40–80 & 11 (14.9\%) & 5 (6.8\%) & 5 (6.8\%) & 53 (71.6\%) \\
Cap\_7 80–240& 7 (11.9\%)  & 0 (0.0\%) & 5 (8.5\%) & 47 (79.7\%) \\
Cap\_8 >240  & 3 (14.3\%)  & 2 (9.5\%) & 3 (14.3\%) & 13 (61.9\%) \\ \hline
\end{tabular}
\label{tab:market-cap}
\end{table}


\begin{table}[htbp]
\centering
\caption{Distribution by Employees Classes}
\begin{tabular}{lrrrr}
\hline
\textbf{Sentiment} & Risk & Risk\_Opportunity & Neutral & Opportunity \\ \hline
Emp\_01 <250      & 14 (34.1\%) & 0 (0.0\%) & 21 (51.2\%) & 6 (14.6\%) \\
Emp\_02 250–1000  & 20 (25.0\%) & 1 (1.2\%) & 49 (61.3\%) & 10 (12.5\%) \\
Emp\_03 1000–5000 & 112 (47.7\%) & 3 (1.3\%) & 105 (44.7\%) & 15 (6.4\%) \\
Emp\_04 5000–10k  & 69 (58.0\%) & 2 (1.7\%) & 45 (37.8\%) & 3 (2.5\%) \\
Emp\_05 10k–40k   & 127 (56.2\%) & 2 (0.9\%) & 84 (37.2\%) & 13 (5.8\%) \\
Emp\_06 40k–100k  & 57 (65.5\%) & 0 (0.0\%) & 25 (28.7\%) & 5 (5.7\%) \\
Emp\_07 100k–250k & 17 (65.4\%) & 0 (0.0\%) & 9 (34.6\%) & 0 (0.0\%) \\
Emp\_08 >250k     & 11 (78.6\%) & 0 (0.0\%) & 1 (7.1\%) & 2 (14.3\%) \\ \hline
\end{tabular}

\vspace{0.5cm}

\begin{tabular}{lrr}
\hline
\textbf{Commitment} & No Commitment & Commitment \\ \hline
Emp\_01 <250      & 8 (19.5\%) & 33 (80.5\%) \\
Emp\_02 250–1000  & 23 (28.7\%) & 57 (71.2\%) \\
Emp\_03 1000–5000 & 32 (13.6\%) & 203 (86.4\%) \\
Emp\_04 5000–10k  & 14 (11.8\%) & 105 (88.2\%) \\
Emp\_05 10k–40k   & 23 (10.2\%) & 203 (89.8\%) \\
Emp\_06 40k–100k  & 5 (5.7\%)   & 82 (94.3\%) \\
Emp\_07 100k–250k & 5 (19.2\%)  & 21 (80.8\%) \\
Emp\_08 >250k     & 2 (14.3\%)  & 12 (85.7\%) \\ \hline
\end{tabular}

\vspace{0.5cm}

\begin{tabular}{lrr}
\hline
\textbf{Specificity} & General & Specific \\ \hline
Emp\_01 <250      & 11 (26.8\%) & 30 (73.2\%) \\
Emp\_02 250–1000  & 27 (33.8\%) & 53 (66.2\%) \\
Emp\_03 1000–5000 & 68 (28.9\%) & 167 (71.1\%) \\
Emp\_04 5000–10k  & 21 (17.6\%) & 98 (82.4\%) \\
Emp\_05 10k–40k   & 43 (19.0\%) & 183 (81.0\%) \\
Emp\_06 40k–100k  & 11 (12.6\%) & 76 (87.4\%) \\
Emp\_07 100k–250k & 3 (11.5\%)  & 23 (88.5\%) \\
Emp\_08 >250k     & 1 (7.1\%)   & 13 (92.9\%) \\ \hline
\end{tabular}

\vspace{0.5cm}

\begin{tabular}{lrrrr}
\hline
\textbf{Net-Zero} & No Reduction & Reduction & Reduction\_Netzero & Netzero \\ \hline
Emp\_01 <250      & 10 (24.4\%) & 1 (2.4\%) & 1 (2.4\%) & 29 (70.7\%) \\
Emp\_02 250–1000  & 22 (27.5\%) & 1 (1.2\%) & 1 (1.2\%) & 56 (70.0\%) \\
Emp\_03 1000–5000 & 48 (20.4\%) & 5 (2.1\%) & 9 (3.8\%) & 173 (73.6\%) \\
Emp\_04 5000–10k  & 19 (16.0\%) & 3 (2.5\%) & 3 (2.5\%) & 94 (79.0\%) \\
Emp\_05 10k–40k   & 38 (16.8\%) & 4 (1.8\%) & 11 (4.9\%) & 173 (76.5\%) \\
Emp\_06 40k–100k  & 4 (4.6\%)   & 4 (4.6\%) & 5 (5.7\%) & 74 (85.1\%) \\
Emp\_07 100k–250k & 2 (7.7\%)   & 2 (7.7\%) & 1 (3.8\%) & 21 (80.8\%) \\
Emp\_08 >250k     & 1 (7.1\%)   & 0 (0.0\%) & 2 (14.3\%) & 11 (78.6\%) \\ \hline
\end{tabular}
\label{tab:employees-classes}
\end{table}


\begin{table}[htbp]
\centering
\caption{Distribution by Sector}
\small
\begin{tabular}{lrrrr}
\hline
\textbf{Sentiment} & Risk & Risk\_Opportunity & Neutral & Opportunity \\ \hline
Communication                  & 20 (66.7\%) & 0 (0.0\%) & 8 (26.7\%) & 2 (6.7\%) \\
Consommation cyclique          & 66 (58.4\%) & 1 (0.9\%) & 39 (34.5\%) & 7 (6.2\%) \\
Consommation non cyclique      & 27 (55.1\%) & 1 (2.0\%) & 19 (38.8\%) & 2 (4.1\%) \\
Finances                       & 27 (36.5\%) & 2 (2.7\%) & 32 (43.2\%) & 13 (17.6\%) \\
Fournisseur                    & 30 (66.7\%) & 0 (0.0\%) & 14 (31.1\%) & 1 (2.2\%) \\
Immobilier                     & 22 (36.7\%) & 0 (0.0\%) & 35 (58.3\%) & 3 (5.0\%) \\
Industrie                      & 88 (60.3\%) & 2 (1.4\%) & 49 (33.6\%) & 7 (4.8\%) \\
Matériaux de base              & 39 (56.5\%) & 0 (0.0\%) & 29 (42.0\%) & 1 (1.4\%) \\
Santé                          & 24 (38.1\%) & 1 (1.6\%) & 31 (49.2\%) & 7 (11.1\%) \\
Technologie de l'information   & 66 (55.0\%) & 1 (0.8\%) & 47 (39.2\%) & 6 (5.0\%) \\
Énergie                        & 18 (30.5\%) & 0 (0.0\%) & 36 (61.0\%) & 5 (8.5\%) \\ \hline
\end{tabular}

\vspace{0.5cm}

\begin{tabular}{lrr}
\hline
\textbf{Commitment} & No Commitment & Commitment \\ \hline
Communication                  & 4 (13.3\%)  & 26 (86.7\%) \\
Consommation cyclique          & 11 (9.7\%)  & 102 (90.3\%) \\
Consommation non cyclique      & 10 (20.4\%) & 39 (79.6\%) \\
Finances                       & 17 (23.0\%) & 57 (77.0\%) \\
Fournisseur                    & 6 (13.3\%)  & 39 (86.7\%) \\
Immobilier                     & 7 (11.7\%)  & 53 (88.3\%) \\
Industrie                      & 14 (9.6\%)  & 132 (90.4\%) \\
Matériaux de base              & 3 (4.3\%)   & 66 (95.7\%) \\
Santé                          & 12 (19.0\%) & 51 (81.0\%) \\
Technologie de l'information   & 13 (10.8\%) & 107 (89.2\%) \\
Énergie                        & 15 (25.4\%) & 44 (74.6\%) \\ \hline
\end{tabular}

\vspace{0.5cm}

\begin{tabular}{lrr}
\hline
\textbf{Specificity} & General & Specific \\ \hline
Communication                  & 6 (20.0\%)  & 24 (80.0\%) \\
Consommation cyclique          & 16 (14.2\%) & 97 (85.8\%) \\
Consommation non cyclique      & 14 (28.6\%) & 35 (71.4\%) \\
Finances                       & 30 (40.5\%) & 44 (59.5\%) \\
Fournisseur                    & 1 (2.2\%)   & 44 (97.8\%) \\
Immobilier                     & 13 (21.7\%) & 47 (78.3\%) \\
Industrie                      & 32 (21.9\%) & 114 (78.1\%) \\
Matériaux de base              & 8 (11.6\%)  & 61 (88.4\%) \\
Santé                          & 20 (31.7\%) & 43 (68.3\%) \\
Technologie de l'information   & 27 (22.5\%) & 93 (77.5\%) \\
Énergie                        & 18 (30.5\%) & 41 (69.5\%) \\ \hline
\end{tabular}

\vspace{0.5cm}

\small
\begin{tabular}{lrrrr}
\hline
\textbf{Net-Zero} & No Reduction & Reduction & Reduction\_Netzero & Netzero \\ \hline
Communication                  & 8 (26.7\%)  & 0 (0.0\%) & 3 (10.0\%) & 19 (63.3\%) \\
Consommation cyclique          & 14 (12.4\%) & 3 (2.7\%) & 6 (5.3\%)  & 90 (79.6\%) \\
Consommation non cyclique      & 9 (18.4\%)  & 1 (2.0\%) & 0 (0.0\%)  & 39 (79.6\%) \\
Finances                       & 27 (36.5\%) & 3 (4.1\%) & 0 (0.0\%)  & 44 (59.5\%) \\
Fournisseur                    & 8 (17.8\%)  & 2 (4.4\%) & 3 (6.7\%)  & 32 (71.1\%) \\
Immobilier                     & 7 (11.7\%)  & 1 (1.7\%) & 3 (5.0\%)  & 49 (81.7\%) \\
Industrie                      & 24 (16.4\%) & 5 (3.4\%) & 8 (5.5\%)  & 109 (74.7\%) \\
Matériaux de base              & 6 (8.7\%)   & 0 (0.0\%) & 3 (4.3\%)  & 60 (87.0\%) \\
Santé                          & 13 (20.6\%) & 0 (0.0\%) & 2 (3.2\%)  & 48 (76.2\%) \\
Technologie de l'information   & 21 (17.5\%) & 4 (3.3\%) & 5 (4.2\%)  & 90 (75.0\%) \\
Énergie                        & 7 (11.9\%)  & 1 (1.7\%) & 0 (0.0\%)  & 51 (86.4\%) \\ \hline
\end{tabular}
\label{tab:sectors}
\end{table}


\begin{table}[htbp]
\centering
\caption{Distribution by Scope EI Classes}
\begin{tabular}{lrrrr}
\hline
\textbf{Sentiment} & Risk & Risk\_Opportunity & Neutral & Opportunity \\ \hline
C1 - Ultra Faible & 106 (42.6\%) & 1 (0.4\%) & 115 (46.2\%) & 27 (10.8\%) \\
C2 - Très Faible  & 57 (43.8\%) & 3 (2.3\%) & 60 (46.2\%) & 10 (7.7\%) \\
C3 - Faible       & 140 (56.9\%) & 3 (1.2\%) & 91 (37.0\%) & 12 (4.9\%) \\
C4 - Moyen        & 44 (58.7\%) & 1 (1.3\%) & 28 (37.3\%) & 2 (2.7\%) \\
C5 - Moyen+       & 31 (60.8\%) & 0 (0.0\%) & 19 (37.3\%) & 1 (2.0\%) \\
C6 - Élevé        & 18 (48.6\%) & 0 (0.0\%) & 17 (45.9\%) & 2 (5.4\%) \\
C7 - Très Élevé   & 27 (77.1\%) & 0 (0.0\%) & 8 (22.9\%)  & 0 (0.0\%) \\
C8 - Extrême      & 4 (80.0\%)  & 0 (0.0\%) & 1 (20.0\%)  & 0 (0.0\%) \\ \hline
\end{tabular}

\vspace{0.5cm}

\begin{tabular}{lrr}
\hline
\textbf{Commitment} & No Commitment & Commitment \\ \hline
C1 - Ultra Faible & 44 (17.7\%) & 205 (82.3\%) \\
C2 - Très Faible  & 14 (10.8\%) & 116 (89.2\%) \\
C3 - Faible       & 34 (13.8\%) & 212 (86.2\%) \\
C4 - Moyen        & 7 (9.3\%)   & 68 (90.7\%) \\
C5 - Moyen+       & 7 (13.7\%)  & 44 (86.3\%) \\
C6 - Élevé        & 5 (13.5\%)  & 32 (86.5\%) \\
C7 - Très Élevé   & 1 (2.9\%)   & 34 (97.1\%) \\
C8 - Extrême      & 0 (0.0\%)   & 5 (100.0\%) \\ \hline
\end{tabular}

\vspace{0.5cm}

\begin{tabular}{lrr}
\hline
\textbf{Specificity} & General & Specific \\ \hline
C1 - Ultra Faible & 84 (33.7\%) & 165 (66.3\%) \\
C2 - Très Faible  & 32 (24.6\%) & 98 (75.4\%) \\
C3 - Faible       & 44 (17.9\%) & 202 (82.1\%) \\
C4 - Moyen        & 12 (16.0\%) & 63 (84.0\%) \\
C5 - Moyen+       & 7 (13.7\%)  & 44 (86.3\%) \\
C6 - Élevé        & 4 (10.8\%)  & 33 (89.2\%) \\
C7 - Très Élevé   & 2 (5.7\%)   & 33 (94.3\%) \\
C8 - Extrême      & 0 (0.0\%)   & 5 (100.0\%) \\ \hline
\end{tabular}

\vspace{0.5cm}

\begin{tabular}{lrrrr}
\hline
\textbf{Net-Zero} & No Reduction & Reduction & Reduction\_Netzero & Netzero \\ \hline
C1 - Ultra Faible & 69 (27.7\%) & 5 (2.0\%) & 8 (3.2\%) & 167 (67.1\%) \\
C2 - Très Faible  & 24 (18.5\%) & 5 (3.8\%) & 3 (2.3\%) & 98 (75.4\%) \\
C3 - Faible       & 30 (12.2\%) & 7 (2.8\%) & 11 (4.5\%) & 198 (80.5\%) \\
C4 - Moyen        & 6 (8.0\%)   & 1 (1.3\%) & 4 (5.3\%) & 64 (85.3\%) \\
C5 - Moyen+       & 3 (5.9\%)   & 0 (0.0\%) & 1 (2.0\%) & 47 (92.2\%) \\
C6 - Élevé        & 4 (10.8\%)  & 1 (2.7\%) & 2 (5.4\%) & 30 (81.1\%) \\
C7 - Très Élevé   & 6 (17.1\%)  & 1 (2.9\%) & 4 (11.4\%) & 24 (68.6\%) \\
C8 - Extrême      & 2 (40.0\%)  & 0 (0.0\%) & 0 (0.0\%) & 3 (60.0\%) \\ \hline
\end{tabular}
\label{tab:scope-ei}
\end{table}


\begin{table}[htbp]
\centering
\caption{Distribution by Scope EJ Classes}
\begin{tabular}{lrrrr}
\hline
\textbf{Sentiment} & Risk & Risk\_Opportunity & Neutral & Opportunity \\ \hline
C1 - Ultra Faible & 58 (36.9\%) & 2 (1.3\%) & 77 (49.0\%) & 20 (12.7\%) \\
C2 - Très Faible  & 65 (45.5\%) & 3 (2.1\%) & 60 (42.0\%) & 15 (10.5\%) \\
C3 - Faible       & 185 (52.0\%) & 3 (0.8\%) & 153 (43.0\%) & 15 (4.2\%) \\
C4 - Moyen        & 58 (63.7\%) & 0 (0.0\%) & 31 (34.1\%) & 2 (2.2\%) \\
C5 - Moyen+       & 29 (70.7\%) & 0 (0.0\%) & 12 (29.3\%) & 0 (0.0\%) \\
C6 - Élevé        & 19 (90.5\%) & 0 (0.0\%) & 2 (9.5\%)  & 0 (0.0\%) \\
C7 - Très Élevé   & 3 (75.0\%)  & 0 (0.0\%) & 1 (25.0\%)  & 0 (0.0\%) \\
C8 - Extrême      & 0 (nan\%)   & 0 (nan\%) & 0 (nan\%)   & 0 (nan\%) \\ \hline
\end{tabular}

\vspace{0.5cm}

\begin{tabular}{lrr}
\hline
\textbf{Commitment} & No Commitment & Commitment \\ \hline
C1 - Ultra Faible & 29 (18.5\%) & 128 (81.5\%) \\
C2 - Très Faible  & 24 (16.8\%) & 119 (83.2\%) \\
C3 - Faible       & 44 (12.4\%) & 312 (87.6\%) \\
C4 - Moyen        & 10 (11.0\%) & 81 (89.0\%) \\
C5 - Moyen+       & 1 (2.4\%)   & 40 (97.6\%) \\
C6 - Élevé        & 0 (0.0\%)   & 21 (100.0\%) \\
C7 - Très Élevé   & 0 (0.0\%)   & 4 (100.0\%) \\
C8 - Extrême      & 0 (nan\%)   & 0 (nan\%) \\ \hline
\end{tabular}

\vspace{0.5cm}

\begin{tabular}{lrr}
\hline
\textbf{Specificity} & General & Specific \\ \hline
C1 - Ultra Faible & 58 (36.9\%) & 99 (63.1\%) \\
C2 - Très Faible  & 40 (28.0\%) & 103 (72.0\%) \\
C3 - Faible       & 65 (18.3\%) & 291 (81.7\%) \\
C4 - Moyen        & 13 (14.3\%) & 78 (85.7\%) \\
C5 - Moyen+       & 4 (9.8\%)   & 37 (90.2\%) \\
C6 - Élevé        & 1 (4.8\%)   & 20 (95.2\%) \\
C7 - Très Élevé   & 1 (25.0\%)  & 3 (75.0\%) \\
C8 - Extrême      & 0 (nan\%)   & 0 (nan\%) \\ \hline
\end{tabular}

\vspace{0.5cm}

\begin{tabular}{lrrrr}
\hline
\textbf{Net-Zero} & No Reduction & Reduction & Reduction\_Netzero & Netzero \\ \hline
C1 - Ultra Faible & 47 (29.9\%) & 3 (1.9\%) & 3 (1.9\%) & 104 (66.2\%) \\
C2 - Très Faible  & 32 (22.4\%) & 3 (2.1\%) & 4 (2.8\%) & 104 (72.7\%) \\
C3 - Faible       & 43 (12.1\%) & 9 (2.5\%) & 14 (3.9\%) & 290 (81.5\%) \\
C4 - Moyen        & 8 (8.8\%)   & 4 (4.4\%) & 5 (5.5\%) & 74 (81.3\%) \\
C5 - Moyen+       & 5 (12.2\%)  & 0 (0.0\%) & 3 (7.3\%) & 33 (80.5\%) \\
C6 - Élevé        & 2 (9.5\%)   & 1 (4.8\%) & 3 (14.3\%) & 15 (71.4\%) \\
C7 - Très Élevé   & 2 (50.0\%)  & 0 (0.0\%) & 1 (25.0\%) & 1 (25.0\%) \\
C8 - Extrême      & 0 (nan\%)   & 0 (nan\%) & 0 (nan\%) & 0 (nan\%) \\ \hline
\end{tabular}
\label{tab:scope-ej}
\end{table}


\begin{table}[htbp]
\centering
\caption{Distribution by Scope EK Classes}
\begin{tabular}{lrrrr}
\hline
\textbf{Sentiment} & Risk & Risk\_Opportunity & Neutral & Opportunity \\ \hline
C1 - Ultra Faible & 14 (42.4\%) & 1 (3.0\%) & 13 (39.4\%) & 5 (15.2\%) \\
C2 - Très Faible  & 11 (42.3\%) & 1 (3.8\%) & 12 (46.2\%) & 2 (7.7\%) \\
C3 - Faible       & 50 (50.0\%) & 0 (0.0\%) & 46 (46.0\%) & 4 (4.0\%) \\
C4 - Moyen        & 30 (44.1\%) & 0 (0.0\%) & 33 (48.5\%) & 5 (7.4\%) \\
C5 - Moyen+       & 46 (59.7\%) & 0 (0.0\%) & 28 (36.4\%) & 3 (3.9\%) \\
C6 - Élevé        & 25 (48.1\%) & 0 (0.0\%) & 27 (51.9\%) & 0 (0.0\%) \\
C7 - Très Élevé   & 57 (75.0\%) & 1 (1.3\%) & 17 (22.4\%) & 1 (1.3\%) \\
C8 - Extrême      & 30 (73.2\%) & 0 (0.0\%) & 10 (24.4\%) & 1 (2.4\%) \\ \hline
\end{tabular}

\vspace{0.5cm}

\begin{tabular}{lrr}
\hline
\textbf{Commitment} & No Commitment & Commitment \\ \hline
C1 - Ultra Faible & 6 (18.2\%) & 27 (81.8\%) \\
C2 - Très Faible  & 6 (23.1\%) & 20 (76.9\%) \\
C3 - Faible       & 10 (10.0\%) & 90 (90.0\%) \\
C4 - Moyen        & 9 (13.2\%) & 59 (86.8\%) \\
C5 - Moyen+       & 6 (7.8\%)  & 71 (92.2\%) \\
C6 - Élevé        & 7 (13.5\%) & 45 (86.5\%) \\
C7 - Très Élevé   & 5 (6.6\%)  & 71 (93.4\%) \\
C8 - Extrême      & 1 (2.4\%)  & 40 (97.6\%) \\ \hline
\end{tabular}

\vspace{0.5cm}

\begin{tabular}{lrr}
\hline
\textbf{Specificity} & General & Specific \\ \hline
C1 - Ultra Faible & 11 (33.3\%) & 22 (66.7\%) \\
C2 - Très Faible  & 3 (11.5\%)  & 23 (88.5\%) \\
C3 - Faible       & 25 (25.0\%) & 75 (75.0\%) \\
C4 - Moyen        & 13 (19.1\%) & 55 (80.9\%) \\
C5 - Moyen+       & 12 (15.6\%) & 65 (84.4\%) \\
C6 - Élevé        & 10 (19.2\%) & 42 (80.8\%) \\
C7 - Très Élevé   & 8 (10.5\%)  & 68 (89.5\%) \\
C8 - Extrême      & 3 (7.3\%)   & 38 (92.7\%) \\ \hline
\end{tabular}

\vspace{0.5cm}

\begin{tabular}{lrrrr}
\hline
\textbf{Net-Zero} & No Reduction & Reduction & Reduction\_Netzero & Netzero \\ \hline
C1 - Ultra Faible & 11 (33.3\%) & 2 (6.1\%) & 1 (3.0\%) & 19 (57.6\%) \\
C2 - Très Faible  & 5 (19.2\%)  & 1 (3.8\%) & 1 (3.8\%) & 19 (73.1\%) \\
C3 - Faible       & 14 (14.0\%) & 3 (3.0\%) & 6 (6.0\%) & 77 (77.0\%) \\
C4 - Moyen        & 9 (13.2\%)  & 3 (4.4\%) & 4 (5.9\%) & 52 (76.5\%) \\
C5 - Moyen+       & 10 (13.0\%) & 0 (0.0\%) & 3 (3.9\%) & 64 (83.1\%) \\
C6 - Élevé        & 5 (9.6\%)   & 0 (0.0\%) & 2 (3.8\%) & 45 (86.5\%) \\
C7 - Très Élevé   & 8 (10.5\%)  & 4 (5.3\%) & 4 (5.3\%) & 60 (78.9\%) \\
C8 - Extrême      & 4 (9.8\%)   & 1 (2.4\%) & 5 (12.2\%) & 31 (75.6\%) \\ \hline
\end{tabular}
\label{tab:scope-ek}
\end{table}


\begin{table}[H]
\centering
\caption{Number of companies by cluster.}
\begin{tabular}{lr}
\hline
\textbf{Cluster} & \textbf{Number of companies} \\ \hline
0 & 183 \\ 
1 & 17 \\ 
2 & 225 \\ 
3 & 111 \\ 
4 & 50 \\ 
5 & 39 \\ 
6 & 43 \\ 
7 & 25 \\ 
8 & 55 \\ 
9 & 80 \\ \hline
\end{tabular}
\label{tab:cluster_counts}
\end{table}

\begin{table}[H]
\centering
\caption{Cluster centroids in original feature space (rounded values).}
\begin{tabular}{rrrrrrr}
\hline
\textbf{Cluster} & \textbf{Sentiment} & \textbf{Commitment} & \textbf{Specificity} & \textbf{Net-zero} & \textbf{Scope~1} & \textbf{Scope~2} \\ \hline
0 & 2.03 & 1.00 & 1.00 & 2.97 & 1.40 & 1.50 \\ 
1 & 2.29 & 0.00 & 0.00 & 2.47 & 1.82 & 2.18 \\ 
2 & 0.00 & 1.00 & 1.00 & 2.93 & 1.25 & 1.52 \\ 
3 & 0.31 & 1.00 & 0.99 & 2.70 & 4.62 & 3.26 \\ 
4 & 2.38 & 0.00 & 0.08 & 0.00 & 0.68 & 0.92 \\ 
5 & 1.26 & 1.00 & 0.00 & 0.10 & 0.72 & 1.18 \\ 
6 & 1.95 & 0.00 & 1.00 & 2.95 & 2.07 & 1.79 \\ 
7 & 0.32 & 0.92 & 0.92 & 2.08 & 2.68 & 6.20 \\ 
8 & 0.62 & 1.00 & 1.00 & 0.22 & 1.31 & 1.38 \\ 
9 & 1.36 & 1.00 & 0.00 & 2.99 & 1.38 & 1.40 \\ \hline
\end{tabular}
\label{tab:cluster_centroids_original}
\end{table}




\begin{figure}[htbp]
  \centering
  \includegraphics[width=0.75\linewidth]{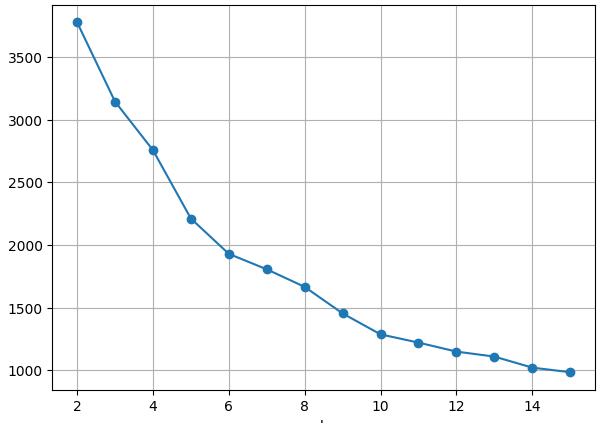}
  \caption{Elbow Method: Inertia vs Number of Clusters}
  \label{fig:elbow}
\end{figure}


\begin{figure}[htbp]
  \centering
  \includegraphics[width=0.8\linewidth]{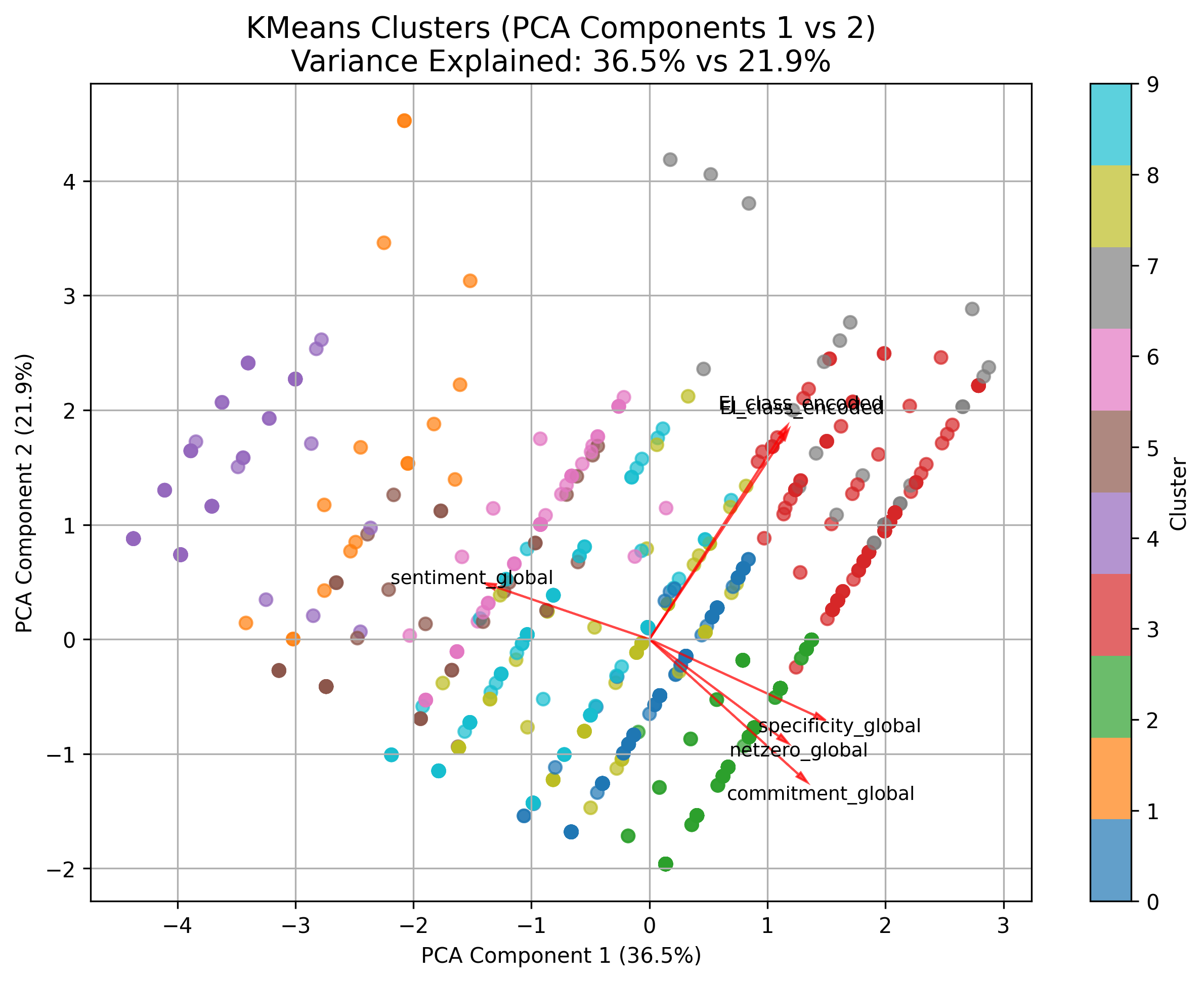}
  \caption{KMeans clusters in PCA space (PC1 vs PC2).}
  \label{fig:pca_pc1_pc2}
\end{figure}

\begin{figure}[htbp]
  \centering
  \includegraphics[width=0.8\linewidth]{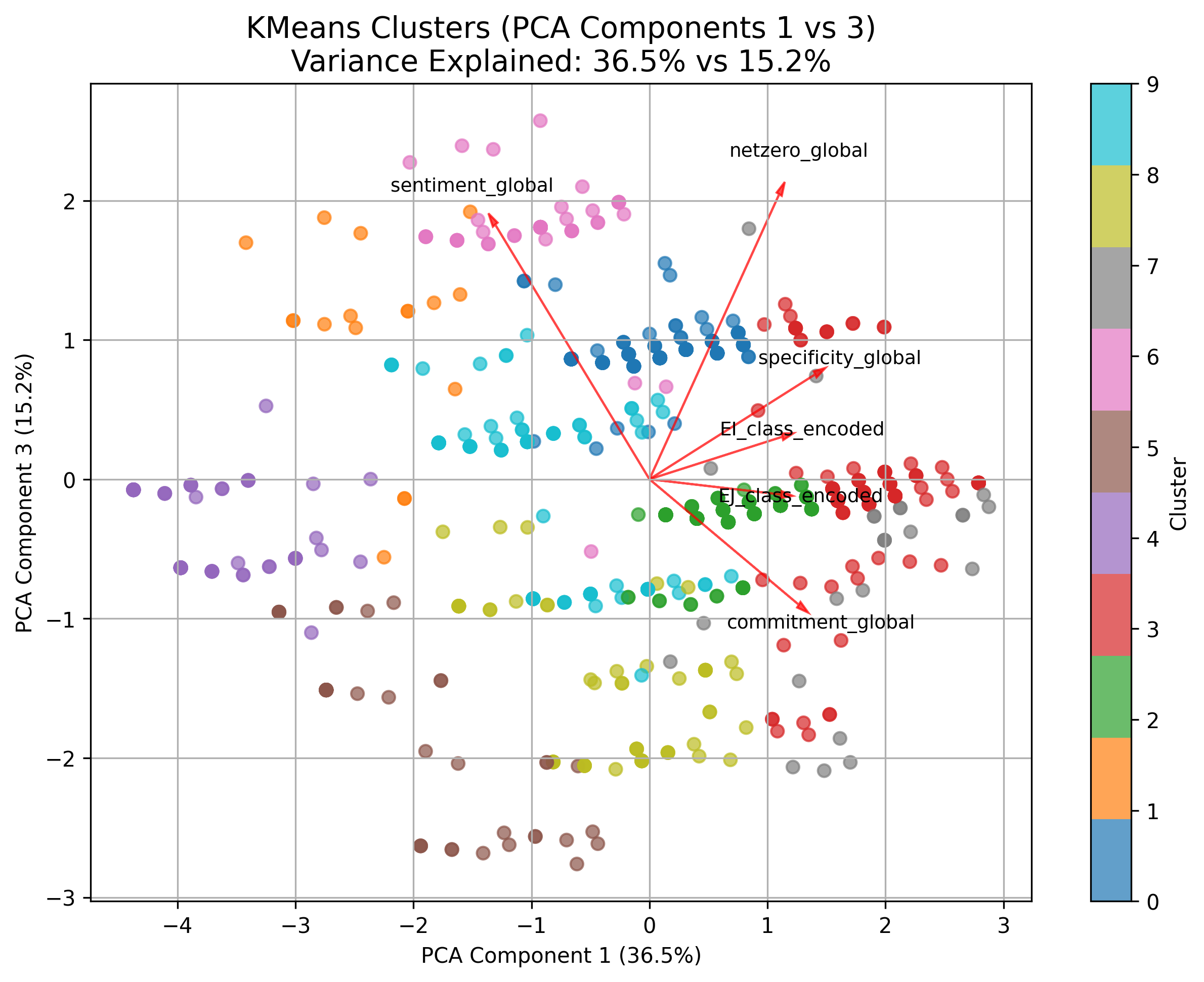}
  \caption{KMeans clusters in PCA space (PC1 vs PC3).}
  \label{fig:pca_pc1_pc2}
\end{figure}
\newpage
\bibliographystyle{plainnat}
\bibliography{bibliography}

\begin{thebibliography}{25}
\providecommand{\natexlab}[1]{#1}
\providecommand{\url}[1]{\texttt{#1}}
\expandafter\ifx\csname urlstyle\endcsname\relax
  \providecommand{\doi}[1]{doi: #1}\else
  \providecommand{\doi}{doi: \begingroup \urlstyle{rm}\Url}\fi

\bibitem[Aerts et~al.(2006)Aerts, Cormier, and Magnan]{AERTS2006299}
Walter Aerts, Denis Cormier, and Michel Magnan.
\newblock Intra-industry imitation in corporate environmental reporting: An international perspective.
\newblock \emph{Journal of Accounting and Public Policy}, 25\penalty0 (3):\penalty0 299--331, 2006.
\newblock ISSN 0278-4254.
\newblock \doi{https://doi.org/10.1016/j.jaccpubpol.2006.03.004}.
\newblock URL \url{https://www.sciencedirect.com/science/article/pii/S0278425406000330}.

\bibitem[Azizyan et~al.(2015)Azizyan, Singh, and Wasserman]{azizyan2015efficient}
Martin Azizyan, Aarti Singh, and Larry Wasserman.
\newblock Efficient sparse clustering of high-dimensional non-spherical gaussian mixtures.
\newblock In \emph{Artificial Intelligence and Statistics}, pages 37--45. PMLR, 2015.

\bibitem[Bakshi and Kothari(2020)]{bakshi2020outlier}
Ainesh Bakshi and Pravesh Kothari.
\newblock Outlier-robust clustering of non-spherical mixtures.
\newblock \emph{arXiv preprint arXiv:2005.02970}, 2020.

\bibitem[Baldini et~al.(2018)Baldini, Maso, Liberatore, Mazzi, and Terzani]{baldini2018role}
Maria Baldini, Lorenzo~Dal Maso, Giovanni Liberatore, Francesco Mazzi, and Simone Terzani.
\newblock Role of country-and firm-level determinants in environmental, social, and governance disclosure.
\newblock \emph{Journal of business ethics}, 150\penalty0 (1):\penalty0 79--98, 2018.

\bibitem[Bebbington(2001)]{bebbington2001sustainable}
Jan Bebbington.
\newblock Sustainable development: a review of the international development, business and accounting literature.
\newblock In \emph{Accounting forum}, volume~25, pages 128--157. Taylor \& Francis, 2001.

\bibitem[Bingler et~al.(2023)Bingler, Kraus, Leippold, and Webersinke]{bingler2023cheaptalk}
Julia Bingler, Mathias Kraus, Markus Leippold, and Nicolas Webersinke.
\newblock How cheap talk in climate disclosures relates to climate initiatives, corporate emissions, and reputation risk.
\newblock Working paper, Available at SSRN 3998435, 2023.

\bibitem[Bingler et~al.(2022)Bingler, Kraus, Leippold, and Webersinke]{bingler2022cheap}
Julia~Anna Bingler, Mathias Kraus, Markus Leippold, and Nicolas Webersinke.
\newblock Cheap talk and cherry-picking: What climatebert has to say on corporate climate risk disclosures.
\newblock \emph{Finance Research Letters}, 47:\penalty0 102776, 2022.

\bibitem[Bini et~al.(2016)Bini, Giunta, and Bellucci]{bini2016put}
Laura Bini, Francesco Giunta, and Marco Bellucci.
\newblock Put your money where your mouth is: The difference between real commitment to sustainability and mere rhetoric.
\newblock \emph{Financial reporting: bilancio, controlli e comunicazione d'azienda: 2, 2016}, pages 5--31, 2016.

\bibitem[Bolton et~al.(2021)Bolton, Reichelstein, Kacperczyk, Leuz, Ormazabal, and Schoenmaker]{bolton2021mandatory}
Patrick Bolton, Stefan~J Reichelstein, Marcin~T Kacperczyk, Christian Leuz, Gaizka Ormazabal, and Dirk Schoenmaker.
\newblock Mandatory corporate carbon disclosures and the path to net zero.
\newblock \emph{Management and Business Review}, 1\penalty0 (3):\penalty0 21--28, 2021.

\bibitem[Cho and Patten(2007)]{CHO2007639}
Charles~H. Cho and Dennis~M. Patten.
\newblock The role of environmental disclosures as tools of legitimacy: A research note.
\newblock \emph{Accounting, Organizations and Society}, 32\penalty0 (7):\penalty0 639--647, 2007.
\newblock ISSN 0361-3682.
\newblock \doi{https://doi.org/10.1016/j.aos.2006.09.009}.
\newblock URL \url{https://www.sciencedirect.com/science/article/pii/S0361368206001036}.

\bibitem[Clarkson et~al.(2008)Clarkson, Li, Richardson, and Vasvari]{CLARKSON2008303}
Peter~M. Clarkson, Yue Li, Gordon~D. Richardson, and Florin~P. Vasvari.
\newblock Revisiting the relation between environmental performance and environmental disclosure: An empirical analysis.
\newblock \emph{Accounting, Organizations and Society}, 33\penalty0 (4):\penalty0 303--327, 2008.
\newblock ISSN 0361-3682.
\newblock \doi{https://doi.org/10.1016/j.aos.2007.05.003}.
\newblock URL \url{https://www.sciencedirect.com/science/article/pii/S0361368207000451}.

\bibitem[Deegan(2002)]{deegan2002introduction}
Craig Deegan.
\newblock Introduction: The legitimising effect of social and environmental disclosures--a theoretical foundation.
\newblock \emph{Accounting, auditing \& accountability journal}, 15\penalty0 (3):\penalty0 282--311, 2002.

\bibitem[Flammer et~al.(2021)Flammer, Toffel, and Viswanathan]{flammer2021shareholder}
Caroline Flammer, Michael~W Toffel, and Kala Viswanathan.
\newblock Shareholder activism and firms' voluntary disclosure of climate change risks.
\newblock \emph{Strategic Management Journal}, 42\penalty0 (10):\penalty0 1850--1879, 2021.

\bibitem[Huang et~al.(2025{\natexlab{a}})Huang, Zhang, Li, Mu, and Wang]{huang2025unmasking}
Qiyu Huang, Yan Zhang, Xiang Li, Xiangning Mu, and Mingyu Wang.
\newblock Unmasking isomorphic behaviors in corporate sustainability: Evidence from esg disclosure and practices in emerging markets.
\newblock \emph{Corporate Social Responsibility and Environmental Management}, 2025{\natexlab{a}}.

\bibitem[Huang et~al.(2025{\natexlab{b}})Huang, Zhang, Li, and Wang]{huang2025imitation}
Qiyu Huang, Yan Zhang, Xiang Li, and Fei Wang.
\newblock Imitation behavior in environmental, social, and governance disclosure: Textual analysis evidence from chinese listed enterprises.
\newblock \emph{Business Ethics, the Environment \& Responsibility}, 34\penalty0 (3):\penalty0 771--793, 2025{\natexlab{b}}.

\bibitem[Ilhan et~al.(2023)Ilhan, Krueger, Sautner, and Starks]{ilhan2023climate}
Emirhan Ilhan, Philipp Krueger, Zacharias Sautner, and Laura~T Starks.
\newblock Climate risk disclosure and institutional investors.
\newblock \emph{The Review of Financial Studies}, 36\penalty0 (7):\penalty0 2617--2650, 2023.

\bibitem[{Independent Scientific Advisory Board (ISAB)}(2018)]{ISAB2018}
{Independent Scientific Advisory Board (ISAB)}.
\newblock Review of the comparative survival study (css) draft 2018 annual report.
\newblock Technical Report ISAB 2018-4, Northwest Power and Conservation Council, Columbia River Basin Indian Tribes, and National Marine Fisheries Service, 851 SW 6th Avenue, Suite 1100, Portland, Oregon 97204, October 2018.
\newblock URL \url{https://www.nwcouncil.org/media/filer_public/c0/f2/c0f2b820-c7bc-414b-bc9d-8b98f0ed23bc/ISAB_2018-4_ReviewCSSdraft2018AnnualReport18Oct.pdf}.
\newblock Also see February 6, 2019 addendum re: CSS Chapter 8. Members: Kurt Fausch, Stan Gregory, William Jaeger, Cynthia Jones, Alec Maule, Peter Moyle, Katherine Myers, Laurel Saito, Steve Schroder, Carl Schwarz, Tom Turner.

\bibitem[Jiang et~al.(2022)Jiang, Hu, Zhang, Chen, Zhong, and Shi]{Jiang2022}
Lu~Jiang, Xiaokang Hu, Gangfeng Zhang, Yanqiang Chen, Honglin Zhong, and Peijun Shi.
\newblock Carbon emission risk and governance.
\newblock \emph{International Journal of Disaster Risk Science}, 13\penalty0 (2):\penalty0 249--260, 2022.
\newblock ISSN 2192-6395.
\newblock \doi{10.1007/s13753-022-00411-8}.
\newblock URL \url{https://doi.org/10.1007/s13753-022-00411-8}.

\bibitem[Lu and Wang(2021)]{LU2021101264}
Jing Lu and Jun Wang.
\newblock Corporate governance, law, culture, environmental performance and csr disclosure: A global perspective.
\newblock \emph{Journal of International Financial Markets, Institutions and Money}, 70:\penalty0 101264, 2021.
\newblock ISSN 1042-4431.
\newblock \doi{https://doi.org/10.1016/j.intfin.2020.101264}.
\newblock URL \url{https://www.sciencedirect.com/science/article/pii/S1042443120301487}.

\bibitem[Manes-Rossi and Nicolo'(2022)]{manes2022exploring}
Francesca Manes-Rossi and Giuseppe Nicolo'.
\newblock Exploring sustainable development goals reporting practices: From symbolic to substantive approaches—evidence from the energy sector.
\newblock \emph{Corporate Social Responsibility and Environmental Management}, 29\penalty0 (5):\penalty0 1799--1815, 2022.

\bibitem[Park et~al.(2023)Park, Nishitani, Kokubu, Freedman, and Weng]{PARK2023135203}
Jin~Dong Park, Kimitaka Nishitani, Katsuhiko Kokubu, Martin Freedman, and Yiting Weng.
\newblock Revisiting sustainability disclosure theories: Evidence from corporate climate change disclosure in the united states and japan.
\newblock \emph{Journal of Cleaner Production}, 382:\penalty0 135203, 2023.
\newblock ISSN 0959-6526.
\newblock \doi{https://doi.org/10.1016/j.jclepro.2022.135203}.
\newblock URL \url{https://www.sciencedirect.com/science/article/pii/S0959652622047771}.

\bibitem[Silva(2021)]{SILVA2021125962}
Samanthi Silva.
\newblock Corporate contributions to the sustainable development goals: An empirical analysis informed by legitimacy theory.
\newblock \emph{Journal of Cleaner Production}, 292:\penalty0 125962, 2021.
\newblock ISSN 0959-6526.
\newblock \doi{https://doi.org/10.1016/j.jclepro.2021.125962}.
\newblock URL \url{https://www.sciencedirect.com/science/article/pii/S0959652621001827}.

\bibitem[Suchman(1995)]{Suchman1995}
Mark~C. Suchman.
\newblock Managing legitimacy: Strategic and institutional approaches.
\newblock \emph{Academy of Management Review}, 20\penalty0 (3):\penalty0 571--610, 1995.
\newblock \doi{10.5465/amr.1995.9508080331}.
\newblock URL \url{https://doi.org/10.5465/amr.1995.9508080331}.

\bibitem[Verrecchia(1983)]{VERRECCHIA1983179}
Robert~E. Verrecchia.
\newblock Discretionary disclosure.
\newblock \emph{Journal of Accounting and Economics}, 5:\penalty0 179--194, 1983.
\newblock ISSN 0165-4101.
\newblock \doi{https://doi.org/10.1016/0165-4101(83)90011-3}.
\newblock URL \url{https://www.sciencedirect.com/science/article/pii/0165410183900113}.

\bibitem[Zou et~al.(2025)Zou, Shi, Chen, Deng, Lei, Zeng, Yang, Tong, Xiao, and Zhou]{zou2025esgreveal}
Yi~Zou, Mengying Shi, Zhongjie Chen, Zhu Deng, ZongXiong Lei, Zihan Zeng, Shiming Yang, Hongxiang Tong, Lei Xiao, and Wenwen Zhou.
\newblock Esgreveal: An llm-based approach for extracting structured data from esg reports.
\newblock \emph{Journal of Cleaner Production}, 489:\penalty0 144572, 2025.

\end{thebibliography}

\end{document}